\documentclass[a4paper]{article}
\usepackage[a4paper, total={5.4in, 8in}]{geometry}
\usepackage{amsmath,amssymb,amsfonts,amsthm}
\usepackage[ruled,vlined,linesnumbered,resetcount,linesnumbered,ruled,vlined]{algorithm2e}
\usepackage{lineno,hyperref}
\usepackage{graphicx}
\usepackage{float}
\usepackage{xcolor}
\usepackage{multirow}
\hypersetup{
    colorlinks,
    linkcolor={red!50!black},
    citecolor={blue!50!black},
    urlcolor={blue!80!black}
}

\title{Solving MaxSAT by Successive Calls to a SAT Solver}
\author{Mohamed El Halaby \\ \small{Department of Mathematics}\\
\small{Faculty of Science}\\
\small{Cairo University}\\
\small{Giza, 12613, Egypt}\\
\href{mailto:halaby@sci.cu.edu.eg}{\small{halaby@sci.cu.edu.eg}}}

\date{}

\newtheorem{exmp}{Example}[section]
\theoremstyle{definition}
\newtheorem{defn}{Definition}[section]

\begin{document}
\maketitle

\begin{abstract} 
The Maximum Satisfiability (MaxSAT) problem is the problem of finding a truth assignment that maximizes the number of satisfied clauses of a given Boolean formula in Conjunctive Normal Form (CNF). Many exact solvers for MaxSAT have been developed during recent years, and many of them were presented in the well-known SAT conference. Algorithms for MaxSAT generally fall into two categories: (1) branch and bound algorithms and (2) algorithms that use successive calls to a SAT solver (SAT-based), which this paper in on. In practical problems, SAT-based algorithms have been shown to be more efficient. This paper provides an experimental investigation to compare the performance of recent SAT-based and branch and bound algorithms on the benchmarks of the MaxSAT Evaluations.
\end{abstract}

\clearpage
\tableofcontents
\clearpage
\listofalgorithms
\clearpage

\section{Introduction and Preliminaries}
A \textit{Boolean variable} $x$ can take one of two possible values 0 (false) or 1 (true). A \textit{literal} $l$ is a variable $x$ or its negation $\neg x$. A \textit{clause} is a disjunction of literals, i.e., $\bigvee_{i=1}^n l_i$. A \textit{CNF formula} is a conjunction of clauses. Formally, a CNF formula $\phi$ composed of $k$ clauses, where each clause $C_i$ is composed of $m_i$ is defined as $F = \bigwedge_{i=1}^k C_i$ where $C_i = \bigvee_{j=1}^{m_i} l_{i,j}$.

In this paper, a set of clauses $\{C_1,C_2,\dots,C_k\}$ is referred to as a Boolean formula. A truth assignment \textit{satisfies} a Boolean formula if it satisfies every clause.

Given a CNF formula $\phi$, the satisfiability problem (SAT) is deciding whether $\phi$ has a satisfying truth assignment (i.e., an assignment to the variables of $\phi$ that satisfies every clause). The \textit{Maximum Satisfiability} (MaxSAT) problem asks for a truth assignment that maximizes the number of satisfied clauses in $\phi$.

Many theoretical and practical problems can be encoded into SAT and MaxSAT such as debugging \cite{safarpour2007improved}, circuits design and scheduling of how an observation satellite captures photos of Earth \cite{vasquez2001logic}, course timetabling \cite{asin2012curriculum,nader2004application,montero2001pso,maric2008timetabling}, software package upgrades \cite{janota2012packup}, routing \cite{xu2003sub,nam2004comparative}, reasoning \cite{sang2007dynamic} and protein structure alignment in bioinformatics \cite{pullan2007protein}.

Let  $\phi=\{(C_1,w_2),\dots,(C_s,w_s)\} \cup \{(C_{s+1},\infty),\dots,(C_{s+h},\infty)\}$ be a CNF formula, where $w_1,\dots,w_s$ are natural numbers. The Weighted Partial MaxSAT problem asks for an assignment that satisfies all $C_{s+1},\dots,C_{s+h}$ (called \textit{hard} clauses) and maximizes the sum of the weights of the satisfied clauses in $C_1,\dots,C_s$ (called \textit{soft} clauses).

In general, exact MaxSAT solvers follow one of two approaches: successively calling a SAT solver (sometimes called the SAT-based approach) and the branch and bound approach. The former converts each MaxSAT problem with different hypothesized maximum weights into multiple SAT problems and uses a SAT solver to solve these SAT problems to determine the actual solution. The SAT-based approach converts the WPMaxSAT problem into a sequence of SAT instances which can be solved using SAT solvers. One way to do this, given an unweighted MaxSAT instance, is to check if there is an assignment that falsifies no clauses. If such an assignment can not be found, we check if there is an assignment that falsifies only one clause. This is repeated and each time we increment the number of clauses that are allowed to be $False$ until the SAT solver returns $True$, meaning that the minimum number of falsified clauses has been determined. Recent comprehensive surveys on SAT-based algorithms can be found in\cite{morgado2013iterative,ansotegui2013sat}.

The second approach utilizes a depth-first branch and bound search in the space of possible assignments. An evaluation function which computes a bound is applied at each search node to determine any pruning opportunity. This paper surveys the satisfiability-based approach and provides an experimental investigation and comparison between the performances of both approaches on sets of benchmarks.

Because of the numerous calls to a SAT solver this approach makes, any improvement to SAT algorithms immediately benefits MaxSAT SAT-based methods. Experimental results from the MaxSAT Evaluations\footnote{Web page: \url{http://www.maxsat.udl.cat}} have shown that SAT-based solvers are more competent to handle large MaxSAT instances from industrial applications than branch and bound methods.

\section{Linear Search Algorithms}
A simple way to solve WPMaxSAT is to augment each soft clause $C_i$ with a new variable (called a blocking variable) $b_i$, then a constraint is added (specified in CNF) saying that the sum of the weights of the falsified soft clauses must be less than a given value $k$. Next, the formula (without the weights) together with the constraint is sent to a SAT solver to check whether or not it is satisfiable. If so, then the cost of the optimal solution is found and the algorithm terminates. Otherwise, $k$ is decreased and the process continues until the SAT solver returns $True$. The algorithm can start searching for the optimal cost from a lower bound $LB$ initialized with the maximum possible cost (i.e. $LB=\sum_{i=1}^{\vert \phi_S \vert} w_i$) and decrease it down to the optimal cost, or it can set $LB = 0$ and increase it up to the optimal cost. Solvers that employ the former approach is called \textit{satisfiability-based} (not to be confused with the name of the general method) solvers, while the ones that follow the latter are called \textit{UNSAT-based} solvers. A cost of 0 means all the soft clauses are satisfied and a cost of means all the soft clauses are falsified.

Algorithm \ref{algo:LinearUNSAT} employs the first method to search for the optimal cost by maintaining (maintaining a lower bound initialized to 0) (line 1). 

{\small
\vspace{0.3in}
\begin{algorithm} [H]
\DontPrintSemicolon % Some LaTeX compilers require you to use \dontprintsemicolon instead
\KwIn{A WPMaxSAT instance $\phi = \phi_S \cup \phi_H$}
\KwOut{A WPMaxSAT solution to $\phi$}

$LB \gets 0$\;

\ForEach{$(C_i,w_i) \in \phi_S$}{
	let $b_i$ be a new blocking variable\;
	$\phi_S \gets \phi_S \setminus \{(C_i,w_i)\} \cup \{(C_i \lor b_i,w_i)\}$
}

\While{$True$}
{
	$(state,I) \gets SAT(\{C \mid (C,w)\in \phi\} \cup CNF(\sum_{i=1}^{\vert \phi_S \vert} w_ib_i \leq LB))$\;
	\If{$state=True$}{
		\Return{$I$}
	}
	$LB \gets UpdateBound(\{w \mid (C,w) \in \phi_S \},LB)$
}
\caption{{LinearUNSAT$(\phi)$} Linear search UNSAT-based algorithm for solving WPMaxSAT.}
\label{algo:LinearUNSAT}
\end{algorithm}
\vspace{0.3in}
}

Next, the algorithm relaxes each soft clause with a new variable in lines 2-4. The formula $\phi$ now contains each soft clause augmented with a new blocking variable. The while loop in lines 5-9 sends the clauses of $\phi$ (without the weights) to a SAT solver (line 6). If the SAT solver returns $True$, then LinearUNSAT terminates returning a solution (lines 7-8). Otherwise, the lower bound is updated and the loop continues until the SAT solver returns $True$. The function $UpdateBound$ in line 9 updates the lower bound either by simply increasing it or by other means that depend on the distribution of the weights of the input formula. Later in this paper we will see how the subset sum problem can be a possible implementation of $UpdateBound$. Note that it could be inefficient if $UpdateBound$ changes $LB$ by one in each iteration. Consider a WPMaxSAT formula with five soft clauses having the weights $1,1,1,1$ and 100. The cost of the optimal solution can not be anything else other than $0,1,2,3,4,100,101,102,103$ and 104. Thus, assigning $LB$ any of the values $5,\dots,99$ is unnecessary and will result in a large number of iterations.

\begin{exmp}
\label{ex:LinearSearch}
Let $\phi=\phi_S \cup \phi_H$, where $\phi_S=\{(x_1,5),(x_2,5),(x_3,10),\\(x_4,5)$, $(x_5,10),(x_6,5),(\neg x_6,10)\}$ and $\phi_H=\{\neg x_1 \lor \neg x_2,\infty),(\neg x_2 \lor \neg x_3,\infty),(\neg x_3 \lor \neg x_4,\infty),(\neg x_4 \lor \neg x_5,\infty),(\neg x_5 \lor \neg x_1,\infty)\}$. If we run LinearUNSAT on $\phi$, the soft clauses will be be relaxed $\{(x_1 \lor b_1,5),(x_2 \lor b_2,5),(x_3 \lor b_3,10),(x_4 \lor b_4,5),(x_5 \lor b_5,10),(x_6 \lor b_6,5),(\neg x_6 \lor b_7,10)\}$ and $LB$ is initialized to 0. The sequence of iterations are
\begin{enumerate}
\item The constraint $CNF(5b_1+5b_2+10b_3+5b_4+10b_5+5b_6+10b_7 \leq 0)$ is included, $state=False$, $LB=5$.

\item The constraint $CNF(5b_1+5b_2+10b_3+5b_4+10b_5+5b_6+10b_7 \leq 5)$ is included, $state=False$, $LB=10$.

\item The constraint $CNF(5b_1+5b_2+10b_3+5b_4+10b_5+5b_6+10b_7 \leq 10)$ is included, $state=False$, $LB=15$.

\item The constraint $CNF(5b_1+5b_2+10b_3+5b_4+10b_5+5b_6+10b_7 \leq 15)$ is included, $state=False$, $LB=20$.

\item The constraint $CNF(5b_1+5b_2+10b_3+5b_4+10b_5+5b_6+10b_7 \leq 20)$ is included, $state=True$. The SAT solver returns the assignment $I=\{x_1=False,x_2=False,x_3=True,x_4=False,x_5=True,x_6=False,b_1=True,b_2=True,b_3=False,b_4=True,b_5=False,b_6=True,b_7=False\}$, which leads to a WPMaxSAT solution if we ignore the values of the $b_i,(1 \leq i \leq 7)$ variables with cost 20.
\end{enumerate}
\end{exmp}

The next algorithm is describes the SAT-based technique. Algorithm \ref{algo:LinearSAT} starts by initializing the upper bound to one plus the the sum of the weights of the soft clauses (line 1).

{\small
\vspace{0.3in}
\begin{algorithm} [H]
\DontPrintSemicolon % Some LaTeX compilers require you to use \dontprintsemicolon instead
\KwIn{A WPMaxSAT instance $\phi = \phi_S \cup \phi_H$}
\KwOut{A WPMaxSAT solution to $\phi$}

$UB \gets 1+\sum_{i=1}^{\vert \phi_S \vert}w_i$\;

\ForEach{$(C_i,w_i) \in \phi_S$}{
	let $b_i$ be a new blocking variable
	$\phi_S \gets \phi_S \setminus \{(C_i,w_i)\} \cup \{(C_i \lor b_i,w_i)\}$
}

\While{$True$}
{
	$(state,I) \gets SAT(\{C \mid (C,w)\in \phi\} \cup CNF(\sum_{i=1}^{\vert \phi_S \vert} w_ib_i \leq UB-1))$\;
	\If{$state=False$}{
		\Return{$lastI$}
	}
	$lastI \gets I$\;
	$UB \gets \sum_{i=1}^{\vert \phi_S \vert}w_i(1-I(C_i\setminus \{b_i\}))$
}
\caption{{LinearSAT$(\phi)$} Linear search SAT-based algorithm for solving WPMaxSAT.}
\label{algo:LinearSAT}
\end{algorithm}
\vspace{0.3in}
}

In each iteration of algorithm \ref{algo:LinearSAT} except the last, the formula is satisfiable. The cost of the optimal solution is found immediately after the transition from satisfiable to unsatisfiable instance. LinearSAT begins by initializing the upper bound to one plus the sum of the weights of the soft clauses (line 1). The while loop (lines 4-8) continues until the formula becomes unsatisfiable (line 6), then the algorithm returns a WPMaxSAT solution and terminates (line 7). As long as the formula is satisfiable, the formula is sent to the SAT solver along with the constraint assuring that the sum of the weights of the falsified soft clauses is less than $UB-1$ (line 5), and the upper bound is updated to the sum of the weights of the soft clauses falsified by the assignment returned by the SAT solver (line 8).

Note that updating the upper bound to $\sum_{i=1}^{\vert \phi_S \vert}w_i(1-I(C_i \setminus \{b_i\}))$ is more efficient than simply decreasing the upper bound by one, because uses less iterations and thus the problem is solved with less SAT calls.

\begin{exmp}
If we run LinearSAT on $\phi$ from the previous example, the soft clauses will be be relaxed $\{(x_1 \lor b_1,5),(x_2 \lor b_2,5),(x_3 \lor b_3,10),(x_4 \lor b_4,5),(x_5 \lor b_5,10),(x_6 \lor b_6,5),(\neg x_6 \lor b_7,10)\}$ and $UB$ is initialized to $1+(5+5+5+5+10+10+10)=51$. The sequence of iterations are
\begin{enumerate}
\item The constraint $CNF(5b_1+5b_2+10b_3+5b_4+10b_5+5b_6+10b_7 \leq 50)$ is included, $state=True$, $I=\{x_1=False,x_2=False,x_3=False,x_4=False,x_5=False,x_6=False,b_1=True,b_2=True,b_3=True,b_4=True,b_5=True,b_6=True,b_7=False\}$, $UB=5+5+10+5+10+5=40$.

\item The constraint $CNF(5b_1+5b_2+10b_3+5b_4+10b_5+5b_6+10b_7 \leq 40-1)$ is included, $state=True$, $I=\{x_1=False,x_2=False,x_3=False,x_4=False,x_5=True,x_6=False,b_1=True,b_2=True,b_3=True,b_4=True,b_5=False,b_6=True,b_7=False\}$, $UB=5+5+10+5+5=30$.

\item The constraint $CNF(5b_1+5b_2+10b_3+5b_4+10b_5+5b_6+10b_7 \leq 30-1)$ is included, $state=True$, $I=\{x_1=False,x_2=False,x_3=True,x_4=False,x_5=True,x_6=False,b_1=True,b_2=True,b_3=False,b_4=True,b_5=False,b_6=True,b_7=False\}$, $UB=5+5+5+5=20$.

\item The constraint $CNF(5b_1+5b_2+10b_3+5b_4+10b_5+5b_6+10b_7 \leq 20-1)$ is included, $state=False$. The assignment from the previous step is indeed a solution to $\phi$ if we ignore the values of the $b_i,(1 \leq i \leq 7)$ variables with cost 20.
\end{enumerate}
\end{exmp}
% % % % % % % % % % % % % % % % % % % % % % % % % % % % % % % % % %
\section{Binary Search-based Algorithms}
The number of iterations linear search algorithms for WPMaxSAT can take is linear in the sum of the weights of the soft clauses. Thus, in the worst case the a linear search WPMaxSAT algorithm can take $\sum_{i=1}^{\vert \phi_S \vert}w_i$ calls to the SAT solver. Since we are searching for a value (the optimal cost) among a set of values (from 0 to $\sum_{i=1}^{\vert \phi_S \vert}w_i$), then binary search can be used, which uses less iterations than linear search. Algorithm \ref{algo:BinarySearch} searches for the cost of the optimal assignment by using binary search.

{\small
\vspace{0.3in}
\begin{algorithm} [H]
\DontPrintSemicolon % Some LaTeX compilers require you to use \dontprintsemicolon instead
\KwIn{A WPMaxSAT instance $\phi = \phi_S \cup \phi_H$}
\KwOut{A WPMaxSAT solution to $\phi$}

$state \gets SAT(\{C_i \mid (C_i,\infty) \in \phi_H \})$\;
\If{$state=False$}{
	\Return{$\emptyset$}
}

$LB \gets -1$\;
$UB \gets 1+\sum_{i=1}^{\vert \phi_S \vert}w_i$\;

\ForEach{$(C_i,w_i) \in \phi_S$}{
	let $b_i$ be a new blocking variable\;
	$\phi_S \gets \phi_S \setminus \{(C_i,w_i)\} \cup \{(C_i \lor b_i,w_i)\}$
}

\While{$LB+1 < UB$}
{
	$mid \gets \lfloor \frac{LB+UB}{2} \rfloor$\;
	$(state,I) \gets SAT(\{C \mid (C,w)\in \phi\} \cup CNF(\sum_{i=1}^{\vert \phi_S \vert} w_ib_i \leq mid))$\;
	\If{$state=True$}{
		$lastI \gets I$\;
		$UB \gets \sum_{i=1}^{\vert \phi_S \vert}w_i(1-I(C_i\setminus \{b_i\}))$
	}
	\Else{
		$LB \gets UpdateBound(\{w_i \mid 1 \leq i \leq \vert \phi_S \vert \},mid)-1$
	}
}
\Return{$lastI$}
\caption{{BinS-WPMaxSAT$(\phi)$} Binary search based algorithm for solving WPMaxSAT.}
\label{algo:BinarySearch}
\end{algorithm}
\vspace{0.3in}
}

BinS-WPMaxSAT begins by checking the satisfiability of the hard clauses (line 1) before beginning the search for the solution. If the SAT solver returns $False$ (line 2), BinS-WPMaxSAT returns the empty assignment and terminates (line 3). The algorithm updates both a lower bound $LB$ and an upper bound $UB$ initialized respectively to -1 and one plus the sum of the weights of the soft clauses (lines 4-5). The soft clauses are augmented with blocking variables (lines 6-8). At each iteration of the main loop (lines 9-16), the middle value ($mid$) is changed to the average of $LB$ and $UB$ and a constraint is added requiring the sum of the weights of the relaxed soft clauses to be less than or equal to the middle value. This clauses describing this constraint are sent to the SAT solver along with the clauses of $\phi$ (line 11). If the SAT solver returns $True$ (line 12), then the cost of the optimal solution is less than $mid$, and $UB$ is updated (line 14). Otherwise, the algorithm looks for the optimal cost above $mid$, and so $LB$ is updated (line 16). The main loop continues until $LB+1=UB$, and the number of iterations BinS-WPMaxSAT executes is proportional to $\log(\sum_{i=1}^{\vert \phi_S \vert} w_i)$ which is a considerably lower complexity than that of linear search methods.

In the following example, $UpdateBound$ assigns $mid+1$ to $LB$.
\begin{exmp}
Consider $\phi$ in example \ref{ex:LinearSearch} with all the weights of the soft clauses set to 1. At the beginning, $LB=-1$, $UB=8$. The following are the sequence of iterations algorithm \ref{algo:BinarySearch} executes.
\begin{enumerate}
\item $mid = \lfloor \frac{8+(-1)}{2} \rfloor=3$, the constraint $CNF(b_1+b_2+b_3+b_4+b_5+b_6+b_7 \leq 3)$ is included, $state=False$, $LB=3$, $UB=8$.

\item $mid = \lfloor \frac{8+3}{2} \rfloor=5$, the constraint $CNF(b_1+b_2+b_3+b_4+b_5+b_6+b_7 \leq 5)$ is included, $state=True$, $I=\{x_1=False,x_2=False,x_3=True,x_4=False,x_5=True,x_6=False,  b_1=True,b_2=True,b_3=False,b_4=True,b_5=False,b_6=True,\\b_7=False\}$, $UB=4$, $LB=3$. The assignment $I$ is indeed an optimal one, falsifying four clauses.
\end{enumerate}
\end{exmp}

It is often stated that a binary search algorithm performs better than linear search. Although this is true most of the time, there are instances for which linear search is faster than binary search. Let $k$ be the sum of the soft clauses falsified by the assignment returned by the SAT solver in the first iteration. If $k$ is indeed the optimal solution, linear search methods would discover this fact in the next iteration, while binary search ones would take $\log k$ iterations to declare $k$ as the optimal cost. In order to benefit from both search methods, An \textit{et al.}\cite{an2011qmaxsat} developed a PMaxSAT algorithm called QMaxSAT (version 0.4) that alternates between linear search and binary search (see algorithm \ref{algo:BinLin}).

\vspace{0.3in}
{\small 
\begin{algorithm} [H]
\DontPrintSemicolon % Some LaTeX compilers require you to use \dontprintsemicolon instead
\KwIn{A WPMaxSAT instance $\phi = \phi_S \cup \phi_H$}
\KwOut{A WPMaxSAT solution to $\phi$}

$state \gets SAT(\{C_i \mid (C_i,\infty) \in \phi_H \})$\;
\If{$state=False$}{
	\Return{$\emptyset$}
}

\ForEach{$(C_i,w_i) \in \phi_S$}{
	let $b_i$ be a new blocking variable\;
	$\phi_S \gets \phi_S \setminus \{(C_i,w_i)\} \cup \{(C_i \lor b_i,w_i)\}$
}

$LB \gets -1$\;
$UB \gets 1+\sum_{i=1}^{\vert \phi_S \vert}w_i$\;
$mode \gets binary$\;

\While{$LB + 1 < UB$}{
	\If{$mode=binary$}{$mid \gets \lfloor \frac{LB+UB}{2} \rfloor$}
	\Else{$mid \gets UB-1$}
	$(state,I) \gets SAT(\{C \mid (C,w)\in \phi \} \cup CNF(\sum_{i=1}^{\vert \phi_S \vert}w_ib_i\leq mid))$\;
	\If{$state=True$}{
		$lastI \gets I$\;
		$UB \gets \sum_{i=1}^{\vert \phi_S \vert}w_i(1-I(C_i\setminus \{b_i\}))$
	}
	\Else{
		\If{$mode=binary$}{$LB \gets UpdateBound(\{w_i \mid 1 \leq i \leq \vert \phi_S \vert \},mid)-1$}
		\Else{$LB \gets mid$}
	}
	\If{$mode=binary$}{$mode \gets linear$}
	\Else{$mode \gets binary$}
}
\Return{$lastI$}

\caption{{BinLin-WPMaxSAT$(\phi)$} Alternating binary and linear searches for solving WPMaxSAT.}
\label{algo:BinLin}
\end{algorithm}
}
\vspace{0.3in}

Algorithm \ref{algo:BinLin} begins by checking that the set of hard clauses is satisfiable (line 1). If not, then the algorithm returns the empty assignment and terminates (line 3). Next, the soft clauses are relaxed (lines 4-6) and the lower and upper bounds are initialized respectively to -1 and one plus the sum of the weights of the soft clauses (lines 7-8). BinLin-WPMaxSAT has two execution modes, binary and linear. The mode of execution is initialized in line 9 to binary search. At each iteration of the main loop (lines 10-27), the SAT solver is called on the clauses of $\phi$ with the constraint $\sum_{i=1}^{\vert \phi_S \vert}w_ib_i$ bounded by the mid point (line 12), if the current mode is binary, or by the upper bound if the mode is linear (line 14). If the formula is satisfiable (line 16), the upper bound is updated. Otherwise, the lower bound is updated to the mid point. At the end of each iteration, the mode of execution is flipped (lines 24-27).

Since the cost of the optimal solution is an integer, it can be represented as an array of bits. Algorithm \ref{algo:BitBased} uses this fact to determine the solution bit by bit. BitBased-WPMaxSAT starts from the most significant bit and at each iteration it moves one bit closer to the least significant bit, at which the optimal cost if found. 

\vspace{0.3in}
{\small 
\begin{algorithm} [H]
\DontPrintSemicolon % Some LaTeX compilers require you to use \dontprintsemicolon instead
\KwIn{A WPMaxSAT instance $\phi = \phi_S \cup \phi_H$}
\KwOut{A WPMaxSAT solution to $\phi$}

$state \gets SAT(\{C_i \mid (C_i,\infty) \in \phi_H \})$\;
\If{$state=False$}{
	\Return{$\emptyset$}
}

\ForEach{$(C_i,w_i) \in \phi_S$}{
	let $b_i$ be a new blocking variable\;
	$\phi_S \gets \phi_S \setminus \{(C_i,w_i)\} \cup \{(C_i \lor b_i,w_i)\}$
}
$k \gets \lfloor \lg(\sum_{i=1}^{\vert \phi_S \vert}w_i) \rfloor$\;
$CurrBit \gets k$\;
$cost \gets 2^k$\;

\While{$CurrBit \geq 0$}{
	$(state,I) \gets SAT(\{C \mid (C,w)\in \phi \} \cup CNF(\sum_{i=1}^{\vert \phi_S \vert} w_ib_i< cost))$\;
	\If{$state = True$}{
		$lastI \gets I$\;
		let $s_0,\dots,s_k \in \{0,1\}$ be constants such that $\sum_{i=1}^{\vert \phi_S \vert}w_i(1-I(C_i\setminus \{b_i\})) = \sum_{j=0}^{k} 2^js_j$ \tcp*[r]{$s_0,\dots,s_k$ are the binary representation of the current cost}
		$CurrBit \gets max(\{j \mid j<CurrBit \text{ and }s_j=1 \} \cup \{-1\})$\;
		\If{$CurrBit \geq 0$}{
			$cost \gets \sum_{j=CurrBit}^{k} 2^js_j$
		}
	}
	\Else{
		$CurrBit \gets CurrBit -1$\;
		$cost \gets cost + 2^{CurrBit}$\;
	}
}
\Return{$lastI$}

\caption{{BitBased-WPMaxSAT$(\phi)$} A bit-based algorithm for solving WPMaxSAT.}
\label{algo:BitBased}
\end{algorithm}
}
\vspace{0.3in}

At the beginning of the algorithm as in the previous ones, the satisfiability of the hard clauses are checked and the soft clauses are relaxed. The sum of the weights of the soft clauses $k$ is an upper bound on the cost and thus it is computed to determine the number of bits needed to represent the optimal solution (line 7). The index of the current bit being considered is initialized to $k$ (line 7), and the value of the solution being constructed is initialized (line 8). The main loop (lines 10-20) terminates when it reached the least significant bit (when $CurrBit = 0$). At each iteration, the SAT solver is called on $\phi$ with constraint saying that the sum of the weights of the falsified soft clauses must be less than $cost$ (line 11). If the SAT solver returns $True$ (line 12), the sum of the weights of the soft clauses falsified by the current assignment is computed and the set of bits needed to represent that number are determined as well (line 14), the index of the current bit is decreased to the next $j<CurrBit$ such that $s_j=1$ (line 15). If such an index does not exist, then $CurrBit$ becomes -1 and in the following iteration the algorithm terminates. On the other hand, if the SAT solver returns $False$, the search continues to the most significant bit by decrementing $CurrBit$ (line 19) and since the optimal cost is greater than the current value of $cost$, it is decreased by $2^{CurrBit}$ (line 20).

\begin{exmp}
Consider $\phi$ from example \ref{ex:LinearSearch} with all the weights of the soft clauses being 1. At the beginning of the algorithm, the soft clauses are relaxed and the formula becomes $\{(x_1 \lor b_1,1),(x_2 \lor b_2,1),(x_3 \lor b_3,1),(x_4 \lor b_4,1),(x_5 \lor b_5,1),(x_6 \lor b_6,1),(\neg x_6 \lor b_7,1)\} \cup \phi_H$. Also, the variables $k$, $CurrBit$ and $cost$ are initialized to 2, 2 and $2^2$ respectively. The following are the iterations BitBased-WPMaxSAT executes.
\begin{enumerate}
\item The constraint $CNF(b_1+b_2+b_3+b_4+b_5+b_6+b_7 < 2^2)$ is included, $state=False$, $CurrBit=1$, $cost = 2^2+2^1=6$.

\item The constraint $CNF(b_1+b_2+b_3+b_4+b_5+b_6+b_7 < 2^2+2^1)$, $state=True$, $I=\{x_1=False,x_2=False,x_3=True,x_4=False,x_5=True,x_6=False,b_1=True,b_2=True,b_3=False,b_4=True,b_5=False,b_6=True,b_7=False\}$, $CurrBit = -1$. 
\end{enumerate}
\end{exmp}
% % % % % % % % % % % % % % % % % % % % % % % % % % % % % % % % % %
\section{Core-guided Algorithms}
	As in the previous method, UNSAT methods use SAT solvers iteratively to solve MaxSAT. Here, the purpose of iterative SAT calls is to identify and relax unsatisfiable formulas (unsatisfiable cores) in a MaxSAT instance. This method was first proposed in 2006 by Fu and Malik in\cite{fu2006solving} (see algorithm \ref{algo:FuMalik}). The algorithms described in this section are
	\begin{enumerate}
	\item Fu and Malik's algorithm\cite{fu2006solving}
	\item WPM1\cite{ansotegui2009solving}
	\item Improved WPM1\cite{ansotegui2012improving}
	\item WPM2\cite{ansotegui2010new}
	\item WMSU1-ROR\cite{heras2011read}
	\item WMSU3\cite{marques2007using}
	\item WMSU4\cite{marques2008algorithms}
	\end{enumerate}

	\begin{defn}[Unsatisfiable core]
		\label{UNSATCore}
		An unsatisfiable core of a CNF formula $\phi$ is a subset of $\phi$ that is unsatisfiable by itself.
	\end{defn}

\begin{defn}[Minimum unsatisfiable core]
A minimum unsatisfiable core contains the smallest number of the original clauses required to still be unsatisfiable. 
\end{defn}
\begin{defn}[Minimal unsatisfiable core]
A minimal unsatisfiable core is an unsatisfiable core such that any proper subset of it is not a core\cite{davies2013postponing}.
\end{defn}
Modern SAT solvers provide the unsatisfiable core as a by-product of the proof of unsatisfiability. The idea in this paradigm is as follows: Given a WPMaxSAT instance $\phi=\{(C_1,w_1),\dots,(C_s,w_s)\} \cup \{(C_{s+1},\infty),\dots,(C_{s+h},\infty)\}$, let $\phi_k$ be a SAT instance that is satisfiable iff $\phi$ has an assignment with cost less than or equal to $k$. To encode $\phi_k$, we can extend every soft clause $C_i$ with a new (auxiliary) variable $b_i$ and add the CNF conversion of the constraint $\sum_{i=1}^s w_ib_i \leq k$. So, we have $$\phi_k = \{(C_i\lor b_i),\dots,(C_s\lor b_s),C_{s+1},\dots,C_{s+h} \} \cup CNF\left( \sum_{i=1}^s w_ib_i \leq k \right)$$

Let $k_{opt}$ be the cost of the optimal assignment of $\phi$. Thus, $\phi_k$ is satisfiable for all $k \geq k_{opt}$, and unsatisfiable for all $k < k_{opt}$, where $k$ may range from 0 to $\sum_{i=1}^s w_i$. Hence, the search for the optimal assignment corresponds to the location of the transition between satisfiable and unsatisfiable $\phi_k$. This encoding guarantees that the all the satisfying assignments (if any) to $\phi_{k_{opt}}$ are the set of optimal assignments to the WPMaxSAT instance $\phi$.

\subsection{Fu and Malik's algorithm}
Fu and Malik implemented two PMaxSAT solvers, ChaffBS (uses binary search to find the optimal cost) and ChaffLS (uses linear search to find the optimal cost) on top of a SAT solver called zChaff\cite{moskewicz2001chaff}. Their PMaxSAT solvers participated in the first and second MaxSAT Evaluations\cite{argelich2008first}. Their method (algorithm \ref{algo:FuMalik} basis for many WPMaxSAT solvers that came later. Notice the input to algorithm \ref{algo:FuMalik} is a PMaxSAT instance since all the weights of the soft clauses are the same.

{\small
\vspace{0.3in}
\begin{algorithm} [H]
\DontPrintSemicolon % Some LaTeX compilers require you to use \dontprintsemicolon instead
\KwIn{$\phi = \{(C_1,1),\dots,(C_s,1),(C_{s+1},\infty),\dots,(C_{s+h},\infty)\}$}
\KwOut{The cost of the optimal assignment to $\phi$}

\If{$SAT(\{C_{s+1},\dots,C_{s+h}\}) = (False,\_)$}{
	\Return{$\infty$}
}
		
$opt \gets 0$ \tcp*[r]{The cost of the optimal solution}
$f \gets 0$ \tcp*[r]{The number of clauses falsified}
\While{$True$}{
$(state,\phi_C) \gets SAT(\{C_i \mid (C_i,w_i) \in \phi\})$ \;
\If{$state = True$}{
	\Return{$opt$}
}
$f \gets f + 1$\;
$B \gets \emptyset$ \;
\ForEach{$C_i \in \phi_C \text{ such that } w_i \neq \infty$}{
let $b_i$ be a new blocking variable\;
$\phi \gets \phi \setminus \{(C_i,1)\} \cup \{(C_i \lor b_i,1)\}$\;
$B \gets B \cup \{i\}$
}
$\phi \gets \phi \cup \{(C,\infty) \mid C \in \sum_{i \in B} b_i=1\}$ \tcp*[l]{Add the cardinality constraint as hard clauses}
$opt \gets opt + 1 $
}
\caption{{Fu\&Malik$(\phi)$} Fu and Malik's algorithm for solving PMaxSAT.}
\label{algo:FuMalik}
\end{algorithm}
\vspace{0.3in}
}

	Fu\&Malik (algorithm \ref{algo:FuMalik}) (also referred to as MSU1) begins by checking if a hard clause is falsified (line 1), and if so it terminates returning the cost $\infty$ (line 2). Next, unsatisfiable cores ($\phi_C$) are identified by iteratively calling a SAT solver on the soft clauses (line 6). If the working formula is satisfiable (line 7), the algorithm halts returning the cost of the optimal assignment (line 8). If not, then the algorithm starts its second phase by relaxing each soft clause in the unsatisfiable core obtained earlier by adding to it a fresh variable, in addition to saving the index of the relaxed clause in $B$ (lines 11-14). Next, the new working formula constraints are added indicating that exactly one of $b_i$ variables should be $True$ (line 15). Finally, the cost is increased by one (line 16) a clause is falsified. This procedure continues until the SAT solver declares the formula satisfiable. 

\subsection{WPM1}
	Ans\'{o}tegui, Bonet and Levy\cite{ansotegui2009solving} extended Fu\& Malik to WPMaxSAT. The resulting algorithm is called WPM1 and is described in algorithm \ref{algo:WPM1}.
	
	\vspace{0.3in}
	{\small
		\begin{algorithm} [H]
			\DontPrintSemicolon % Some LaTeX compilers require you to use \dontprintsemicolon instead
			\KwIn{A WPMaxSAT instance $\phi = \{(H_1,\infty),\dots,(H_h,\infty)\} \cup \{(S_1,w_1),\dots,(S_s,w_s)\}$}
			\KwOut{The optimal cost of the WPMaxSAT solution}
			\If{$SAT(\{H_i \mid 1 \leq i \leq h \})=False$}{
				\Return{$\infty$}
			}
			$cost \gets 0$\;
			\While{$True$}{
				$(state,\phi_C) \gets SAT(\{C_i \mid (C_i,w_i) \in \phi\})$\;
				\If{$state=True$}{
					\Return{$cost$}
				}
				$BV \gets \emptyset$\;
				$w_{min} \gets min\{w_i \mid C_i \in \phi_C \text{ and } w_i \neq \infty \}$\;  \tcp*[l]{Compute the minimum weight of all the soft clauses in $\phi_C$}
				\ForEach{$C_i \in \phi_C$}{
					\If{$w_i \neq \infty$}{
						Let $b_i$ be a new blocking variable\;
						$\phi \gets \phi \setminus \{(C_i,w_i)\} \cup \{(C_i,w_i-w_{min})\} \cup \{(C_i \lor b_i,w_{min})\}$\;
						$BV \gets BV \cup \{b_i\}$\;
					}
				}
				\If{$BV = \emptyset$}{
					\Return{$False$} \tcp*[l]{$\phi$ is unsatisfiable}
				}
				\Else{
					$\phi \gets \phi \cup CNF\left(\sum_{b \in BV} b=1 \right)$ \tcp*[l]{Add the cardinality constraint as hard clauses}
				}
				$cost \gets cost + w_{min}$
			}
			
			\caption{{WPM1$(\phi)$} The WPM1 algorithm for WPMaxSAT.}
			\label{algo:WPM1}
		\end{algorithm}
	}
	\vspace{0.3in}
	
Just as in Fu\&Malik, algorithm \ref{algo:WPM1} calls a SAT solver iteratively with the working formula, but without the weights (line 5). After the SAT solver returns an unsatisfiable core, the algorithm terminates if the core contains hard clauses and if it does not, then the algorithm computes the minimum weight of the clauses in the core, $w_{min}$ (line 9). Next, the working formula is transformed by duplicating the core (line 13) with one copy having the clauses associated with the original weight minus the minimum weight and a second copy having having the clauses augmented with blocking variables with the original weight. Finally, the cardinality constraint on the blocking variable is added as hard clauses (line 18) and the cost is increased by the minimum weight (line 19).

WPM1 uses blocking variables in an efficient way. That is, if an unsatisfiable core, $\phi_C=\{C_1,\dots,C_k \}$, appears $l$ times, all the copies get the same set of blocking variables. This is possible because the two formulae $\phi_1 = \phi \setminus \phi_C \cup \{C_1 \lor b_i, \dots, C_i \lor b_i \mid C_i \in \phi_C\} \cup CNF\left(\sum_{i=1}^k b_i = 1\right)$ and $\phi_2 = \phi \setminus \phi_C \cup  \{C_i \lor b_i^1, \dots, C_i \lor b_i^l \mid C_i \in \phi_C\} \cup CNF\left(\sum_{i=1}^k b_i^1 = 1\right) \cup \dots \cup CNF\left(\sum_{i=1}^k b_i^l = 1\right)$ are MaxSAT equivalent, meaning that the minimum number of unsatisfiable clause of $\phi_1$ and $\phi_2$ is the same. However, the algorithm does not avoid using more than one blocking variable per clause. This disadvantage is eliminated by WMSU3 (described later).

\begin{exmp}
Consider $\phi=\{(x_1,1),(x_2,2),(x_3,3),(\neg x_1 \lor \neg x_2,\infty),\\(x_1 \lor \neg x_3,\infty),(x_2 \lor \neg x_3,\infty)\}$. In the following, $b_i^j$ is the relaxation variable added to clause $C_i$ at the $j$th iteration. A possible execution sequence of the algorithm is:
\begin{enumerate}
\item $state=False$, $\phi_C=\{(\neg x_3),(\neg x_1 \lor \neg x_2),(x_1 \lor \neg x_3),(x_2 \lor \neg x_3)\}$, $w_{min}=3$, $\phi=\{(x_1,1),(x_2,2),(x_3\lor b_3^1,3),$, $(\neg x_1 \lor \neg x_2,\infty),(x_1 \lor \neg x_3,\infty),(x_2 \lor \neg x_3,\infty),(b_3^1=1,\infty)\}$.

\item $state=False$, $\phi_C=\{(x_1),(x_2),(\neg x_1 \lor \neg x_2)\}$, $w_{min}=1$, $\phi=\{(x_1 \lor b_1^2),(x_2,1),(x_2 \lor b_2^2),(x_3 \lor b_3^1),(\neg x_1 \lor \neg x_2,\infty),$$ $$(x_1 \lor \neg x_3,\infty),\\(x_2 \lor \neg x_3,\infty),(b_3^1=1,\infty),(b_1^2 + b_2^2=1,\infty)$.

\item $state=True$, $A=\{x_1=0,x_2=1,x_3=0\}$ is \textbf{an} optimal assignment with $$\sum_{\substack{
   C_i \text{ is soft } \\
   A \text{ satisfies } C_i
  }}
 w_i = 2
$$
\end{enumerate}
If the SAT solver returns a different unsatisfiable core in the first iteration, a different execution sequence is going to take place.
\end{exmp}

\subsection{Improved WPM1}
In 2012, Ans\'{o}tegui, Bonet and Levy presented a modification to WPM1 (algorithm \ref{algo:WPM1})\cite{ansotegui2012improving}. In WPM1, the clauses of the core are duplicated after computing their minimum weight $w_{min}$. Each clause $C_i$ in the core, the $(C_i,w_i-w_{min}) \text{ and } (C_i \lor b_i,w_{min})$ are added to the working formula and $(C_i,w_i)$ is removed. This process of duplication can be inefficient because a clause with weight $w$ can be converted into $w$ copies with weight 1. The authors provided the following example to illustrate this issue: consider $\phi=\{(x_1,1),(x_2,w),(\neg x_2,\\\infty)\}$. If the SAT solver always includes the first clause in the identified core, the working formula after the first iteration will be $\{(x_1 \lor b_1^1,1),(x_2 \lor b_2^1,1),(x_2,w-1),(\neg x_2,\infty),(b_1^1+b_2^1=1,\infty)\}$. If at each iteration $i$, the SAT solver includes the first clause and with $\{(x_2,w-i+1),(\neg x_2,\infty)\}$ in the unsatisfiable core, then after $i$ iterations the formula would be $\{(x_1 \lor b_1^1 \lor \dots \lor b_1^i,1),(x_2 \lor b_2*1,1),\dots,(x_2 \lor b_2^i,1),(x_2,w-i),(\neg x_2,\infty),(b_1^1 + b_2^1=1,\infty),\dots,(b_1^i + b_2^i=1,\infty) \}$. In this case, WPM1 would need $w$ iterations to solve the problem. 
		
		\vspace{0.3in}
		{\small
			\begin{algorithm} [H]
				\DontPrintSemicolon
				\KwIn{A WPMaxSAT instance $\phi=\{(C_1,w_1),\dots,(C_m,w_m),(C_{m+1},\infty),\dots,(C_{m+m'},w_{m+m'}) \}$}
				\KwOut{The cost of the optimal WPMaxSAT solution to $\phi$}
				
				\If{$SAT(\{C_i \mid w_i=\infty \})=(False,\_)$}{
					\Return{$\infty$} \tcp*[l]{$cost=\infty$ if the hard clauses can not be satisfied}
				}
				$cost \gets 0$\;
				$w_{max} \gets max\{w_i \mid (C_i,w_i) \in \phi \text{ and } w_i < w_{max} \}$  \tcp*[l]{Initialize $w_{max}$ to the largest weight smaller than $\infty$}
				
				\While{$True$}{
					$(state,\phi_C) \gets SAT\left( \{C_i \mid (C_i,w_i) \in \phi \text{ and } w_i \geq w_{max} \}\right) $\;
					\If{$state = True \text{ and } w_{max} = 0$}{
						\Return{$cost$}
					}
					\Else{
						\If{$state = True$}  {
							$w_{nax}=max\{w_i \mid (C_i,w_i)\in \phi \text{ and } w_i < w_{max} \}$}
						
						\Else{
							$BV \gets \emptyset$ \tcp*[l]{Set of blocking variables of the unsatisfiable core}
							$w_{min} \gets min\{w_i \mid C_i \in \phi_C \text{ and } w_i \neq \infty \}$  \tcp*[l]{Minimum weight of soft clauses in the unsatisfiable core}
							\ForEach{$C_i \in \phi_C$}{
								\If{$w_i \neq \infty$}{
									Let $b$ be a new variable
									$\phi \gets \phi \setminus \{(C_i,w_i)\} \cup \{(C_i,w_i-w_{min}),(C_i \lor b,w_{min})\}$\;
									$BV \gets BV \cup \{b\}$\;
								}
							}
							$\phi \gets \phi \cup \{(C,\infty) \mid C \in CNF(\sum_{b \in BV} b=1) \}$  \tcp*[l]{The cardinality constraint is added as hard clauses}
							$cost \gets cost + w_{min}$\;
						}
					}}
					\caption{{ImprovedWPM1$(\phi)$} The stratified approach for WPM1 algorithm.}
					\label{algo:ImprovedWPM1}
				\end{algorithm}}
				\vspace{0.3in}
				
Algorithm \ref{algo:ImprovedWPM1} overcomes this problem by utilizing a stratified approach. The aim is to restrict the clauses sent to the SAT solver to force it to concentrate on those with higher weights, which leads the SAT solver to return unsatisfiable cores with clauses having larger weights. Cores with clauses having larger weight are better because they contribute to increasing the cost faster. Clauses with lower weights are used after the SAT solver returns $True$. The algorithm starts by initializing $w_{max}$ to the largest weight smaller than $\infty$, then in line 6 only the clauses having weight greater than or equal to $w_{max}$ are sent to the SAT solver. The algorithm terminates if the SAT solver returns $True$ and $w_{max}$ is zero (lines 7-8), but if $w_{max}$ is not zero and the formula is satisfiable then $w_{max}$ is decreased to the largest weight smaller than $w_{max}$ (lines 10-11). When the SAT solver returns $False$, the algorithm proceeds as the regular WPM1.
				
A potential problem with the stratified  approach is that in the worst case the algorithm could use more calls to the SAT solver than the regular WPM1. This is because there is no contribution made to the cost when the SAT solver returns $True$ and at the same time $w_{max} > 0$. The authors apply the \textit{diversity heuristic} which decreases $w_{max}$ faster when there is a big variety of distinct weights and assigns $w_{max}$ to the next value of $w_i$ when there is a low diversity among the weights.

\subsection{WPM2}
In 2007, Marques-Silva and Planes\cite{marques2007using} discussed important properties of Fu\&Malik that were not mentioned in\cite{fu2006solving}. If $m$ is the number of clauses in the input formula, they proved that the algorithm performs $O(m)$ iterations and the number of relaxation variables used in the worst case is $O(m^2)$. Marques-Silva and Planes also tried to improve the work of Fu and Malik. Fu\&Malik use the pairwise encoding\cite{gent2002arc} for the constraints on the relaxation variables, which use a quadratic number of clauses. This becomes impractical when solving real-world instances. Instead, Marques-Silva and Planes suggested several other encodings all of which are linear in the number of variables in the constraint\cite{warners1998linear,sinz2005towards,een2006translating,gent2002arc}.
	
Another drawback of Fu\&Malik is that there can be several blocking variables associated with a given clause. This is due to the fact that a clause $C$ can participate in more than one unsatisfiable core. Each time $C$ is a part of a computed unsatisfiable core, a new blocking variable is added to $C$. Although the number of blocking variables per clause is possibly large (but still linear), at most one of these variables can be used to prevent the clause from participating in an unsatisfiable core. A simple solution to reduce the search space associated with blocking variables is to require that at most one of the blocking variables belonging to a given clause can be assigned $True$. For a clause $C_i$, let $b_{i,j},(1\leq j\leq t_i)$ be the blocking variables associated with $C_i$. The condition $\sum_{j=1}^{t_i} b_{i,j} \leq 1$ assures that at most one of the blocking variables of $C_i$ is assigned $True$. This is useful when executing a large number of iterations, and many clauses are involved in a significant number of unsatisfiable cores. The resulting algorithm  that incorporated these improvements is called MSU2.

Ans\'{o}tegui, Bonet and Levy also developed an algorithm for WPMaxSAT in 2010, called WPM2\cite{ansotegui2010new}, where every soft clause $C_i$ is extended with a unique fresh blocking variable $b_i$. Note that a SAT solver will assign $b_i$ $True$ if $C_i$ is $False$. At every iteration, the algorithm modifies two sets of at-most and at-least constraints on the blocking variables, called $AL$ and $AM$ respectively. The algorithm relies of the notion of \textit{covers}. 
\begin{defn}[Cover]
\label{DefCover}
Given a set of cores $L$, its set of covers $Covers(L)$ is defined as the minimal partition of $\{1,\dots,m\}$ such that for every $A \in L$ and $B \in Covers(L)$, if $A \cap B \neq \emptyset$, then $A \subseteq B$.
\end{defn} 

{\small
		\vspace{0.3in}
		\begin{algorithm} [H]
			\DontPrintSemicolon
			\KwIn{A WPMaxSAT instance $\phi=\{(C_1,w_1),\dots,(C_m,w_m),(C_{m+1},\infty),\dots,(C_{m+m'},\infty) \}$}
			\KwOut{The optimal WPMaxSAT solution to $\phi$}
			\If{$SAT(\{C_i \in \phi \mid w_i=\infty  \})=(False,\_)$}{
				\Return{$\infty$}
			}
			$\phi^e \gets \{C_1 \lor b_1,\dots,C_m \lor b_m,C_{m+1},\dots,C_{m+m'}\}$\;
			$Covers \gets \{\{1\},\dots,\{m\}\}$\;
			$AL \gets \emptyset$\;
			$AM \gets \{w_1b_1 \leq 0,\dots,w_mb_m \leq 0\}$\;
			\While{$True$}{
				$(state,\phi_C,I) \gets SAT\left(\phi^e \cup CNF(AL \cup AM) \right)$\;
				\If{$state=True$}{
					\Return{$I$}
					%\Return{$\sum \{k' \mid \sum_{i\in B'} w_ib_i \leq k' \in AM\}$}
				}
				Remove the hard clauses from $\phi_C$\;
				\If{$\phi_C = \emptyset$}{
					\Return{$\emptyset$} \tcp*[r]{$\phi$ has no solution}
				}
				$A \gets \emptyset$\;
				\ForEach{$C_i \lor b_i \in \phi_C $}{
					$A \gets A \cup \{i\}$\;
				}
				$RC \gets \{B \in Covers \mid B \cap A \neq \emptyset \}$\;
				$B \gets \bigcup_{B' \in RC} B'$\;
				$k \gets NewBound(AL,B)$\;
				$Covers \gets Covers \setminus RC \cup B$\;
				$AL \gets AL \cup \{\sum_{i \in B} w_ib_i \geq k \}$\;
				$AM \gets AM \setminus \{\sum_{i \in B'} w_ib_i \leq k' \mid B' \in RC \} \cup \{\sum_{i \in B} w_ib_i \leq k \}$
			}
			\caption{{WPM2$(\phi)$} The WPM2 algorithm for WPMaxSAT}
			\label{algo:WPM2}
		\end{algorithm}
		\vspace{0.3in}
		}
		
		The constraints in $AL$ give lower bounds on the optimal cost of $\phi$, while the ones in $AM$ ensure that all solutions of the set $AM \cup AL$ are the solutions of $AL$ of minimal cost. This in turn ensures that any solution of $\phi^e \cup CNF(AL \cup AM)$ (if there is any) is an optimal assignment of $\phi$. 
		
		The authors use the following definition of \textit{cores} and introduced a new notion called \textit{covers} to show how $AM$ is computed given $AL$.

\begin{defn}[Core]
\label{DefCore}
A core is a set of indices $A$ such that $$\left( \sum_{i \in A} w_ib_i \geq\\ k\right)  \in AL$$. The function $Core\left(\sum_{i \in A} w_ib_i \geq k \right)$ returns the core $A$, and $Cores(AL)$ returns $\{Core(al) \mid al \in AL\}$.
\end{defn}

\begin{defn}[Disjoint cores]
\label{DisjointCore}
Let $U=\{U_1,\dots,U_k\}$ be a set of unsatisfiable cores, each with a set of blocking variables $B_i,(1 \leq i \leq k)$. A core $U_i \in U$ is disjoint if for all $U_j \in U$ we have $(R_i \cap R_j =\emptyset \text{ and } i \neq j)$
\end{defn}

		Given a set of $AL$ constraints, $AM$ is the set of at-most constraints $\sum_{i \in A} w_ib_i \leq k$ such that $A \in Cover(Cores(AL))$ and $k$ is the solution minimizing $\sum_{i \in A} w_ib_i$ subject to $AL$ and $b_i \in \{True,False\}$. At the beginning,  $AL=\{w_1b_1 \geq 0,\dots,w_mb_m \geq 0 \}$ and the corresponding $AM=\{w_1b_1 \leq 0,\dots,w_mb_m \leq 0 \}$ which ensures that the solution to $AL \cup AM$ is $b_1=False,\dots,b_m=False$. At every iteration, when an unsatisfiable core $\phi_C$ is identified by the SAT solver, the set of indices of soft clauses in $\phi_C$ $A \subseteq \{1,\dots,m\}$ is computed, which is also called a core. Next, the set of covers $RC = \{B' \in Covers \mid B' \cap A \neq \emptyset\}$ that intersect with $A$ is computed, as well as their union $B = \bigcup_{B' \in RC} B'$. The new set of covers is $Covers = Covers \setminus RC \cup B$. The set of at-least constraints $AL$ is enlarged by adding a new constraint $\sum_{i \in B} w_ib_i \geq NewBound(AL,B)$, where $NewBound(AL,B)$ correspond to minimize $\sum_{i \in A} w_ib_i$ subject to the set of constraints $\{\sum_{w_ib_i \geq k} \} \cup AL$ where $k=1+\sum \{k' \mid \sum_{i \in A'}w_ib_i \leq k' \in AM \text{ and } A' \subseteq A  \}$. Given $AL$ and $B$, the computation of $NewBound$ can be difficult since it can be reduced to the subset sum problem in the following way: given $\{w_1,\dots,w_n \}$ and $k$, minimize $\sum_{j=1}^n w_jx_j$ subject to $\sum_{j=1}^n w_jx_j > k$ and $x_j \in \{0,1\}$. This is equivalent to $NewBound(AL,B)$, where the weights are $w_j$, $B=\{1,\dots,n \}$ and $AL=\{\sum_{j=1}^n w_jx_j \geq k\}$. In the authors' implementation, $NewBound$ is computed by algorithm \ref{algo:NewBoundWPM2}. 
		
		{\small
		\vspace{0.3in}
		\begin{algorithm} [H]
			\DontPrintSemicolon
			$k \gets \sum\left\lbrace k' \mid \sum_{i \in B'} w_ib_i \leq k' \in AM \text{ and } B' \subseteq B \right\rbrace $\;
			\Repeat{$SAT\left(CNF\left(AL \cup \{\sum_{i \in B} w_ib_i=k \} \right)  \right) $}{
				$k \gets SubsetSum(\{w_i \mid i \in B\},k)$\;
			}
			\Return{$k$}
			\caption{{NewBound$(AL,B)$}}
			\label{algo:NewBoundWPM2}
		\end{algorithm}
		\vspace{0.3in}
		}
		
		The $SubsetSum$ function (called in line 3) is an optimization version of the decision subset sum problem. It returns the largest integer $d \leq k$ such that there is a subset of $\{w_i \mid i\in B\}$ that sums to $d$.
		
\begin{exmp}
Consider $\phi$ in example $\ref{ex:LinearSearch}$ with all the weights of the soft clauses set to 1. Before the main loop of algorithm \ref{algo:WPM2}, we have $\phi^e=\{(x_1\lor b_1),(x_2\lor b_2),(x_3\lor b_3),(x_4\lor b_4),(x_5\lor b_5),(x_6\lor b_6),(\neg x_6\lor b_7)\} \cup \phi_H$, $Covers = \{ \{1\},\{2\},\{3\},\{4\},\{5\},\{6\},\{7\} \}$, $AL=\emptyset$, $AM=\{b_1 \leq 0,b_2 \leq 0,b_3 \leq 0,b_4 \leq 0,b_5 \leq 0,b_6 \leq 0,b_7 \leq 0\}$. The following are the iterations the algorithm executes. The soft clauses in the core $\phi_C$ are denoted by $Soft(\phi_C)$.
\begin{enumerate}
\item $state=False$, $Soft(\phi_C)=\{(x_6 \lor b_6),(\neg x_6 \lor b_7)\}$, $A=\{6,7\}$, $RC=\{ \{6\},\{7\} \}$, $B=\{6,7\}$, $k=1$, $Covers=\{ \{1\},\{2\},\{3\},\\ \{4\},\{5\},\{6,7\} \}$, $AL=\{b_6+b_7 \geq 1 \}$, $AM=\{b_1 \leq 0,b_2 \leq 0,b_3 \leq 0,b_4 \leq 0,b_5 \leq 0,b_6+b_7 \leq 1\}$.

\item $state=False$, $Soft(\phi_C)=\{(x_1),(x_2)\}$, $A=\{1,2\}$, $RC=\{\{1\}, \{2\} \}$, $B=\{1,2\}$, $k=1$, $Covers=\{ \{1,2\},\{3\},\{4\},\{5\},\\ \{6,7\} \}$, $AL=\{b_6+b_7 \geq 1,b_1+b_2 \geq 1 \}$, $AM=\{b_3 \leq 0,b_4 \leq 0,b_5 \leq 0,b_6+b_7 \leq 1, b_1+b_2 \leq 1 \}$.

\item $state=False$, $Soft(\phi_C)=\{(x_3),(x_4)\}$, $A=\{3,4\}$, $RC=\{\{3\}, \{4\}\}$, $B=\{3,4\}$, $k=1$, $Covers=\{\{1,2\},\{3,4\},\{5\},\{6,7\}\}$, $AL=\{b_6+b_7\geq 1, b_1+b_2 \geq 1, b_3+b_4 \geq 1\}$, $AM=\{b_1+b_2 \leq 1, b_5 \leq 0, b_6+b_7 \leq 1, b_3+b_4 \leq 1 \}$.

\item $state=False$, $Soft(\phi_C)=\{(x_1),(x_2),(x_3),(x_4),(x_5)\}$, $A=\{1,2,3,4,5\}$, $RC=\{ \{1,2\}, \{3,4\}, \{5\} \}$, $B=\{1,2,3,4,5\}$, $k=3$, $Covers=\{\{6,7\},\{1,\\2,3,4,5\}\}$, $AL=\{b_6+b_7\geq 1, b_1+b_2 \geq 1, b_3+b_4 \geq 1, b_1+b_2+b_3+b_4+b_5 \geq 3\}$, $AM=\{b_1+b_2 \leq 1,b_1+b_2+b_3+b_4+b_5 \leq 3\}$.

\item $state=True$, $I=\{x_1=False,x_2=False,x_3=True,x_4=False,x_5=True,x_6=False,b_1=True,b_2=True,b_3=False,b_4=True,b_5=False,b_6=True,b_7=False\}$.
\end{enumerate}
\end{exmp}
		
		To sum up, the WPM2 algorithm groups the identified cores in covers, which are a decomposition of the cores into disjoint sets. Constraints are added so that the relaxation variables in each cover relax a particular weight of clauses $k$, which is changed to the next largest value the weights of the clauses can sum up to. Computing the next $k$ can be expensive since it relies on the subset sum problem, which is NP-hard. 

In\cite{ansotegui2013improving}, Ans\'{o}tegui \textit{et at.} invented three improvements to WPM2. First, they applied the stratification technique\cite{ansotegui2012improving}. Second, they introduced a new criteria to decide when soft clauses can be hardened. Finally, they showed that by focusing search on solving to optimality subformulae of the original WPMaxSAT instance, they efficiency of WPM2 is increased. This allows to combine the strength of exploiting the information extracted from unsatisfiable cores and other optimization approaches. By solving these smaller optimization problems the authors obtained the most significant boost in their new WPM2 version.

\subsection{WMSU1-ROR}
		WMSU1-ROR\cite{heras2011read} is a modification of WPM1. It attempts to avoid adding blocking variables by applying MaxSAT resolution to the clauses of the unsatisfiable core. Given an unsatisfiable core $\phi_C$, a resolution refutation (a contradiction obtained by performing resolution) is  calculated by a specialized tool. As much of this refutation as possible is copied by applying MaxSAT resolution steps to the working formula. If the transformation derived the empty clause, it means that the core is trivial and the sequence of calls to the SAT solver can continue without adding any relaxation variables for this step. Otherwise, the transformed core is relaxed as in WPM1. The classical resolution rule can not be applied in MaxSAT because it does not preserve the equivalence among weighted formulae. The MaxSAT resolution rule used in WMSU1-ROR is called Max-RES and is described in\cite{larrosa2008logical}. The following definition extends the resolution rule from SAT to WMaxSAT.
		\begin{defn}[WPMaxSAT resolution]
		$\{(x \lor A,u),(\neg x \lor B,w)\} \equiv \{(A \lor B,m),(x \lor A,u \circleddash m),(\neg x \lor B,w \circleddash m),(x \lor A \lor \neg B,m),(\neg x \lor \neg A \lor B,m)\}$, where $A$ and $B$ are disjunctions and $\circleddash$ is defined on weights $u,w \in \{0,\dots,\top\}$, such that $u \geq w$, as $$u \circleddash w = \left\{
		\begin{array}{ll}
		u-w & u \neq \top \\
		\top & u = \top
		\end{array}
		\right.$$
		and $m=min(u,w)$. The clauses $(x \lor A,u)$ and $(\neg x \lor B,w)$ are called the \textit{clashing clauses}, $(A \lor B,m)$ is called the \textit{resolvent}, $(x \lor A,u \circleddash m)$ and $(\neg x \lor B,w \circleddash m)$ are called \textit{posterior clashing} clauses, $(x \lor A \lor \neg B,m)$ and $(\neg x \lor \neg A \lor B,m)$ are the \textit{compensation} clauses (which are added to recover an equivalent MaxSAT formula).
		\end{defn}
		
		For example, if Max-RES is applied on $\{(x \lor y,3),(\neg x \lor y \lor z,4)\}$ with $\top > 4$, we obtain $\{(y \lor y \lor z,3),(x \lor y,3 \circleddash 3),(\neg x \lor y \lor z,4 \circleddash 3),(x \lor y \lor \neg(y \lor z),3),(\neg x \lor \neg y \lor y \lor z,3)\}$. The first and fourth clauses can be simplified by observing that $(A \lor C \lor \neg (C \lor B),u) \equiv (A \lor C \lor \neg B,u)$. The second and fifth clauses can be deleted since the former has weight zero and the latter is a tautology. De Morgan's laws can not be applied on MaxSAT instance for not preserving the equivalence among instances\cite{larrosa2008logical}. The following rule can be applied instead $(A \lor \neg(l \lor C),w) \equiv \{(A \lor \neg C),(A \lor \neg l \lor C,w)\}$. A resolution proof is an ordered set $R=\{C_i=(C_{i'} \bowtie C_{i''}),C_{i+1}=(C_{i'+1} \bowtie C_{i''+1}),\dots,C_{i+k}=(C_{i'+k} \bowtie C_{i''+k}) \}$, where $(C_i,w_i)=(C_{i'},w_{i'}) \bowtie (C_{i''},w_{i''})$ is the
		the resolution step $i$ of a resolution proof, $(C_i,w_i)$ is the resolvent and $(C_{i'},w_{i'})$ and $(C_{i''},w_{i''})$ are the clashing clauses. The set of compensation clauses will be denoted $[(C_{i'},w_{i'}) \bowtie (C_{i''},w_{i''})]$.
		
		The ROR approach is captured in lines 12-22 in algorithm \ref{algo:ROR}. WMSU1-ROR handles WPMaxSAT formulae the same way as\cite{ansotegui2009solving}. It maintains a working formula $\phi_W$ and a lower bound $LB$. The resolution proof $R_C$ is obtained in line 12 and MaxSAT resolution is applied (lines 14-21) for each read-once step. In detail, the weights of the clashing clauses $(C_{i'},w_{i'})$ and $(C_{i''},w_{i''})$ are decreased by the minimum weight of the clauses in the unsatisfiable core $\phi_C$ (lines 15-16). If the clashing clauses are soft, they are deleted from $\phi_C$ (lines 17-18) and if their resolvent is not $\square$, it is added to $\phi_C$ (lines 21-22). On the other hand, if the clashing clauses are hard, they are kept in the core because they could be used in a different resolution step. Lastly, the compensation and clashing clauses are added to $\phi_W$ (lines 19-20). 
		
		\vspace{0.3in}
		{\small
			\begin{algorithm} [H]
				\DontPrintSemicolon
				\KwIn{A WPMaxSAT instance $\phi=\{(C_1,w_1),\dots,(C_m,w_m),(C_{m+1},\infty),\dots,(C_{m+m'},w_{m+m'}) \}$}
				\KwOut{The cost of the optimal solution to $\phi$}
				\If{$SAT(\{C_i \mid w_i=\infty \})=False$}{
					\Return{$\infty$}
				}
				$\phi_W \gets \phi$\;
				$LB \gets 0$\;
				\While{$True$}{
					$(state,\phi_C) \gets SAT(\phi_W)$\;
					\If{$state = True$}{
						\Return{$LB$}
					}
					$\phi_W \gets \phi_W \setminus \phi_C$\;
					$m \gets min\left( \{w \mid (C,w) \in \phi_C \text{ and } w < \top  \}\right) $\;
					$LB \gets LB + m$\;
					\tcp*[l]{Beginning of read-once resolution}
					$R_C \gets  GetProof(\phi_C)$\;
					\ForEach{$(C_i,w_i) = (C_{i'},w_{i'}) \bowtie (C_{i''},w_{i''}) \in R_C$}{
						\If{$ROR((C_i,w_i),R_C)$}{
							$(C_{i'},w_{i'}) \gets (C_{i'},w_{i'} \circleddash m)$\;
							$(C_{i''},w_{i''}) \gets (C_{i''},w_{i''} \circleddash m)$\;
							\If{$w_{i'} < \top \text{ and } w_{i''} < \top$}{
								$\phi_C \gets \phi_C \setminus \{(C_{i'},w_{i'}),(C_{i''},w_{i''})\}$\;
							}
							$\phi_W \gets \phi_W \cup \{(C_{i'},w_{i'}),(C_{i''},w_{i''})\}$\;
							$\phi_W \gets \phi_W \cup \{[(C_{i'},w_{i'}) \bowtie (C_{i''},w_{i''})]\}$\;
							\If{$C_i \neq \square$}{
								$\phi_C \gets \phi_C \cup \{(C_i,m)\}$
							}
						}
					}
					\tcp*[l]{End of read-once resolution}
					$B \gets \emptyset$\;
					\ForEach{$(C_i,w_i) \in \{(C,w) \mid (C,w) \in \phi_C \text{ and } w < \top \}$}{
						Let $b$ be a new relaxation variable
						$B \gets B \cup \{b\}$
						$\phi_C \gets \phi_C \cup \{(C \lor b,m)\}$
						\If{$w > m $}{
							$(C,w) \gets (C,w \circleddash m)$
						}
						\Else{
							$\phi_C \gets \phi_C \setminus \{(C,w)\}$
						}
					}
					$\phi_c \gets \phi_C \cup CNF\left( \sum_{b \in B} b = 1 \right) $\;
					$\phi_W \gets \phi_W \cup \phi_C$
				}
				
				\caption{WMSU1-ROR($\phi$)}
				\label{algo:ROR}
			\end{algorithm}
		}
		\vspace{0.3in}
		
		$Hard((C_i,w_i),R)$ (algorithm \ref{algo:Hard-ROR}) returns $True$ if $(C_i,w_i)$ is a hard clause and all its ancestors are hard, otherwise it returns $False$. $Input((C_i,w_i),R)$ (called in line 1) returns $True$ if $(C_i,w_i)$ is not a resolvent of any step in $R$ (i.e., an original clause), otherwise it returns $False$. $ancestors((C_i,w_i),R)$ (called in line 5) returns the pair of clauses $(C_{i'},w_{i'})$ and $(C_{i''},w_{i''})$ from which $(C_i,w_i)$ was derived as dictated by $R$.
		
		{\small
		\vspace{0.3in}
		\begin{algorithm} [H]
			\DontPrintSemicolon
			\KwIn{A proof $R$}
			\KwOut{$True$ if $(C_i,w_i)$ is hard, or $False$ otherwise}
			
			\If{$Input((C_i,w_i),R)$ and $w_i = \top$}{
				\Return{$True$}
			}
			\If{$Input((C_i,w_i),R)$ and $w_i \neq \top$}{
				\Return{$False$}
			}
			$\{(C_{i'},w_{i'}),(C_{i''},w_{i''})\} \gets ancestors((C_i,w_i),R)$\;
			
			\Return{$Hard((C_{i'},w_{i'}),R) \text{ and } Hard((C_{i''},w_{i''}),R)$}
			\caption{{Hard$\left( (C_i,w_i),R\right)$} Determines if a clause is hard or not}
			%\caption{(Hard$\left( (C_i,w_i),R\right) $) Determines if a clause is hard or not}
			\label{algo:Hard-ROR}
		\end{algorithm}
		\vspace{0.2in}
		}
		
		The function $ROR$ (algorithm \ref{algo:ROR-Help}) returns $True$ if $(C_i,w_i)$ is hard or if it and all of its soft ancestors have been used at most once in the resolution proof $R$. If $(C_k,w_k)=(C_{k'},w_{k'}) \bowtie (C_{k''},w_{k''})$, where $(C_k,w_k)$ is the last resolvent in a resolution proof $R$. The entire proof is read-once if $ROR((C_k,w_k),R)$ returns $True$. In this case (when the last step is ROR), the resolvent of that step is $(\square,m)$. If this situation occurs, the algorithm does not need to augment clauses with relaxation variables or cardinality constraints, which improves upon the original algorithm. 
		
		\vspace{0.2in}
		{\small
		\begin{algorithm} [H]
			\DontPrintSemicolon
			\KwIn{A proof $R$}
			\KwOut{$True$ if $(C_i,w_i)$ is hard or if its ancestors are used exactly once, $False$ otherwise}
			
			\If{$Hard((C_i,w_i),R)$}{
				\Return{$True$}
			}
			\If{$Input((C_i,w_i),R)$ and $Used((C_i,w_i),R)=1$}{
				\Return{$True$}
			}
			\If{$Used((C_i,w_i),R)>1$}{
				\Return{$False$}
			}
			$\{(C_{i'},w_{i'}),(C_{i''},w_{i''})\} \gets ancestors((C_i,w_i),R)$\;
			
			\Return{$ROR((C_{i'},w_{i'}),R) \text{ and } ROR((C_{i''},w_{i''}),R)$}
			
			\caption{{ROR$\left( (C_i,w_i),R\right)$} Determines if a clause is hard or not or if its ancestors are used at most once}
			%\caption{ROR$\left( (C_i,w_i),R\right) $ Determines if a clause is hard or not}
			\label{algo:ROR-Help}
		\end{algorithm}
		}
		\vspace{0.3in}
		
		The problem with this approach (applying Max-RES instead of adding blocking variables and cardinality constraints) is that when soft clauses with weights greater than zero are resolved more than once, MaxSAT resolution does not ensure to produce resolvents with weights greater than zero. For this technique to work, the authors restrict the application of  resolution to the case where each clause is used at most once, which is referred to as \textit{read-once resolution} (\textit{ROR}). Unfortunately, ROR can not generate resolution proofs for some unsatisfiable clauses\cite{iwama1995intractability}.

\subsection{WMSU3}
WMSU3 is a WPMaxSAT algorithm that adds a single blocking variable per soft clause, thus limiting the number of variables in the formula sent to the SAT solver in each iteration. 

{\small
\vspace{0.3in}
\begin{algorithm} [H]
\DontPrintSemicolon
\KwIn{A WPMaxSAT instance $\phi=\phi_S \cup \phi_H$}
\KwOut{The cost of the optimal WPMaxSAT solution to $\phi$}

\If{$SAT(\{C \mid (C,\infty)\in \phi_H\})=False$}{
	\Return{$\infty$}
}

$B \gets \emptyset$ \tcp*[r]{Set of blocking variables}
$\phi_W \gets \phi$ \tcp*[r]{Working formula initialized to $\phi$}
$LB \gets 0$ \tcp*[r]{Lower bound initialized to 0}

\While{$True$}{
	$(state,\phi_C) \gets SAT(\{C \mid (C,w)\in \phi_W \} \cup CNF(\sum_{b_i \in B} w_ib_i \leq LB))$\;
	\If{$state = True$}{
		\Return{$LB$}
	}
	\ForEach{$(C_i,w_i) \in \phi_C \cap \phi_S$}{
		\If{$w \neq \infty$}{
			$B \gets B \cup \{b_i\}$\;
			$\phi_W \gets \phi_W \setminus \{(C_i,w_i)\} \cup \{(C_i \lor b_i,w_i)\}$
		}
	}
	$LB \gets UpdateBound(\{w_i\mid b_i \in B \},LB)$
}

\caption{{WMSU3$(\phi)$} The WMSU3 algorithm for WPMaxSAT.}
\label{algo:WMSU3}
\end{algorithm}
\vspace{0.3in}
}

Algorithm \ref{algo:WMSU3} begins by initializing the set of blocking variables that will be augmented later to $\emptyset$ (line 3), the working formula to $\phi$ (line 4) and the lower bound to zero (line 5). MSU3 then loops over unsatisfiable working formulae $\phi_W$ (while loop in lines 6-13) until it finds a satisfiable one in line 8. At each iteration, when an unsatisfiable core is returned by the SAT solver, the algorithm adds one blocking variable to each soft clause that has not been augmented with a blocking variable yet (line 13), unlike WPMaxSAT algorithms discussed previously such as WPM1 (algorithm \ref{algo:WPM1}). Indeed, at most one blocking variable is added to each clause because if at iteration $i$ $C_i$ was blocked by $b_i$, then at iteration $i+1$ the clause $C_i \lor b_i$ will not be in $\phi_C \cap \phi_S$. The function $UpdateBound$ in line 14 updates the lower bound LB, either by simply incrementing it or by the subset sum problem as in\cite{ansotegui2010new}. The following example illustrates how the algorithm works. 

% This is an example from a paper 
\begin{exmp}
Let $\phi=\{(x_1,1),(x_2,3),(x_3,1)\} \cup \{(\neg x_1 \lor \neg x_2,\infty),(\neg x_2 \lor \neg x_3,\infty)\}$.
\begin{enumerate}
\item $state = False$, $\phi_C = \{(x_1),(x_2),(\neg x_1 \lor \neg x_2)\}$, $\phi_C \cap \phi_S = \{(x_1),(x_2)\}$, $\phi_W=\{(x_1 \lor b_1,1),(x_2 \lor b_2,2),(x_3,1),(\neg x_1 \lor \neg x_2,\infty),(\neg x_2 \lor \neg x_3,\infty)\}$, $LB = 1$. 

\item The constraint $CNF(b_1+3b_2 \leq 1)$ is included and satisfying it implies that $b_2$ must be falsified, and thus $CNF(b_1+3b_2 \leq 1)$ is replaced by $(\neg b_2)$. $state=False$, $\phi_C = \{(x_2 \lor b_2),(x_3),(\neg x_2 \lor \neg x_3),(\neg b_2)\}$, $\phi_C \cap \phi_S =\{(x_3)\}$, $\phi_W=\{(x_1 \lor b_1,1),(x_2 \lor b_2,2),(x_3 \lor b_3,1),(\neg x_1 \lor \neg x_2,\infty),(\neg x_2 \lor \neg x_3,\infty)\}$, $LB=2$. As in the previous iteration, satisfying the constraint $b_1+3b_2+b_3 \leq 2$ implies $b_2$ must be falsified.

\item The constraint $CNF(b_1+3b_2+b_3 \leq 2)$ is included, $state=True$ and the assignment $I=\{x_1=False,x_2=True,x_3=False,b_1=True,b_2=False,b_3=True\}$ indeed satisfies $\phi_W$ of the last iteration. By ignoring the values of the blocking variables, $I$ is indeed an optimal assignment for $\phi$. It falsifies the soft clauses $(x_1,1)$ and $(x_3,1)$ and satisfies $(x_2,3)$.
\end{enumerate}
\end{exmp} 

\subsection{WMSU4}
Like WMSU3, WMSU4\cite{marques2008algorithms} (algorithm \ref{algo:WMSU4}) adds at most one blocking variable to each soft clause. Thought, it maintains an upper bound ($UB$) as well as a lower bound ($LB$). If the current working formula is satisfiable (line 9), $UB$ is changed to the sum of the weights of the falsified clauses by the solution ($I$) returned from the SAT solver. On the other hand, if the working formula is unsatisfiable, the SAT solver returns an unsatisfiable core, and the algorithm adds a blocking variable to each clause that has not yet been relaxed in that core. If all the soft clauses in the unsatisfiable core have been relaxed (line 16), then the algorithm updates the lower bound (line 17) and exists the main loop. The following example illustrates how the algorithm works.

\vspace{0.3in}
\begin{algorithm} [H]
\DontPrintSemicolon
\KwIn{A WPMaxSAT instance $\phi=\phi_S \cup \phi_H$}
\KwOut{The cost of the optimal WPMaxSAT solution to $\phi$}

\If{$SAT(\{C \mid (C,\infty) \in \phi_H \})=False$}{\Return{$\infty$}}

$B \gets \emptyset$ \tcp*[r]{Set of blocking variables}
$\phi_W \gets \phi$ \tcp*[r]{Working formula initialized to $\phi$}
$LB \gets -1$ \tcp*[r]{Lower bound initialized to 0}
$UB \gets 1 + \sum_{i=1}^{\vert \phi_S \vert} w_i$ \tcp*[r]{Upper bound initialized to the sum of the weights of the soft clauses plus one}
\While{$UB > LB + 1$}{
	$(state,\phi_C,I) \gets SAT(\{C \mid (C,w)\in \phi_W \} \cup CNF(\sum_{b_i \in B} w_ib_i \leq UB-1))$\;
	\If{$state = True$}{
		$UB \gets \sum_{b_i \in B} w_i(1-I(C_i \setminus b_i))$  \tcp*[r]{Update $UB$ to the sum of the weights of the falsified clauses without the blocking variables}
	}
	\Else{
	\ForEach{$(C_i,w_i) \in \phi_C \cap \phi_S$}{
		\If{$w \neq \infty$}{
			$B' \gets B' \cup \{b_i\}$\;
			$\phi_W \gets \phi_W \setminus \{(C_i,w_i)\} \cup \{(C_i \lor b_i,w_i)\}$
		}
	}
	\If{$B' = \emptyset$}{$LB \gets UB - 1$}
			\Else{
				$B \gets B \cup B'$\;
				$LB \gets UpdateBound(\{w_i\mid b_i \in B \},LB)$
			}
	}
}
\Return{$UB$}

\caption{{WMSU4$(\phi)$} The WMSU4 algorithm for WPMaxSAT.}
\label{algo:WMSU4}
\end{algorithm}
\vspace{0.3in}

\begin{exmp}
Let $\phi=\phi_S \cup \phi_H$, where $\phi_S = \{(x_1,1),(x_2,1),(x_3,1),\\(x_4,1)\}$ and $\phi_H= \{(\neg x_1 \lor \neg x_2,\infty),(\neg x_1 \lor \neg x_3,\infty),(\neg x_1 \lor \neg x_4,\infty),(\neg x_2 \lor \neg x_3 \lor \neg x_4,\infty)\}$. Before the first iteration of the while loop, we have $LB = -1$, $UB = 1 + (1+1+1+1)=5$ and $\phi_W = \phi$.
\begin{enumerate}
\item $state = False$, $\phi_C \cap \phi_S = \{(x_2),(x_3),(x_4)\}$, $LB = 0$, $\phi_W=\{(x_1,1),(x_2 \lor b_2,1),(x_3 \lor b_3,1),(x_4 \lor b_4,1)\} \cup \phi_H$.

\item The constraint $CNF(b_2 + b_3 + b_4 \leq 5-1)$ is included, $state=True$, $I=\{x_1=True,x_2=False,x_3=False,x_4=False,b_2=True,b_3=True,b_4=True\}$, $UB=3$.

\item The constraint $CNF(b_2 + b_3 + b_4 \leq 3-1)$ is included, $state=False$, $\phi_C \cap \phi_S = \{(x_1,1),(x_2 \lor b_2,1),(x_3 \lor b_3,1),(x-4 \lor b_4,1)\}$, $LB=1$, $\phi_W = \{(x_1\lor b_1),(x_2 \lor b_2,1),(x_3 \lor b_3,1),(x-4 \lor b_4,1)\} \cup \phi_H$.

\item The constraint $CNF(b_1 + b_2 + b_3 + b_4 \leq 2)$ is included, $state=SAT$, $I=\{x_1=False,x_2=False,x_3=True,x_4=True,b_1=True,b_2=True,b_3=False,b_4=False\}$, $UB=2$. The cost of the optimal assignment is indeed 2 (since $(x_1,1)$ and $(x_2,1)$ are falsified) by $I$.
\end{enumerate}
\end{exmp}

\section{Core-guided Binary Search Algorithms}
Core-guided binary search algorithms are similar to binary search algorithms described in the first section, except that they do not augment all the soft clauses with blocking variables before the beginning of the main loop.
Heras, Morgado and Marques-Silva proposed this technique in\cite{heras2011core} (see algorithm \ref{algo:CoreGuided-BS}). 

\vspace{0.3in}
\begin{algorithm} [H]
\DontPrintSemicolon
\KwIn{A WPMaxSAT instance $\phi=\phi_S \cup \phi_H$}
\KwOut{The cost of the optimal WPMaxSAT solution to $\phi$}

$state \gets SAT(\{C_i \mid (C_i,\infty) \in \phi_H \})$\;
\If{$state=False$}{
	\Return{$\emptyset$}
}

$\phi_W \gets \phi$\;
$LB \gets -1$\;
$UB \gets 1+\sum_{i=1}^{\vert \phi_S \vert}w_i$\;
$B \gets \emptyset$

\While{$LB+1<UB$}{
	$mid \gets \lfloor \frac{LB+UB}{2} \rfloor$\;
	$(state,\phi_C,I) \gets SAT(\{C \mid (C,w) \in \phi_W\} \cup CNF(\sum_{b_i \in B} w_ib_i \leq mid))$\;
	\If{$state=True$}{
		$UB \gets \sum_{i=1}^{\vert \phi_S \vert} w_i(1-I(C_i\setminus \{b_i\}))$\;
		$lastI \gets I$\;
	}
	\Else{
		\If{$\phi_C \cap \phi_S=\emptyset$}{
			$LB \gets UpdateBound(\{w_i \mid b_i \in B\},mid)-1$
		}
		\Else{
			\ForEach{$(C_i,w_i) \in \phi_C \cap \phi_S$}{
			let $b_i$ be a new blocking variable\;
			$B \gets B \cup \{b_i\}$\;
			$\phi_W \gets \phi_W \setminus \{(C_i,w_i)\} \cup \{(C_i \lor b_i,w_i)\}$
		}}
	}
	\Return{$lastI$}
}

\caption{{CoreGuided-BS$(\phi)$} Core-guided binary search algorithm for solving WPMaxSAT.}
\label{algo:CoreGuided-BS}
\end{algorithm}
\vspace{0.3in}

Similar to other algorithms, CoreGuided-BS begins by checking the satisfiability of the hard clauses (lines 1-3). Then it initializes the lower bound (line 4), the upper bound (line 5) and the set of blocking variables (line 6) respectively to -1, one plus the sum of the weights of the soft clauses and $\emptyset$. At each iteration of the main loop (lines 7-21) a SAT solver is called on the working formula with a constraint ensuring that the sum of the weights of the relaxed soft clauses is less than or equal the middle value (line 9). If the formula is satisfiable (line 10), the upper bound is updated to the sum of the falsified soft clauses by the current assignment (line 11). Otherwise, if all the soft clauses have been relaxed (line 14), then the lower bound is updated (line 15), and if not, non-relaxed sot clauses belonging to the core are relaxed (lines 17-19). The main loop continues as long as $LB+1 < UB$.

\begin{exmp}
Consider $\phi$ in example \ref{ex:LinearSearch} with all the weights of the soft clauses set to 1. At the beginning of the algorithm $LB=-1$, $UB=8$, $B=\emptyset$ and $\phi_H$ is satisfiable. The following are the iterations the algorithm executes.
\begin{enumerate}
\item $mid = \lfloor \frac{-1+8}{2} \rfloor=3$. Since $B = \emptyset$, no constraint is included. $state=False$, $\phi_C \cap \phi_S = \{(x_6),(\neg x_6)\}$, $B=\{b_6,b_7\}$. $\phi = \{(x_1,1),(x_2,1),(x_3,1),\\(x_4,1),(x_5,1),(x_6 \lor b_6,1),(\neg x_6 \lor b_7,1)\} \cup \phi_H$.

\item $mid=3$, the constraint $CNF(b_6+b_7 \leq 3)$ is included. $state=False$, $\phi_C \cap \phi_S = \{(x_1),(x_2)\}$, $B=\{b_1,b_2,b_6,b_7\}$, $\phi=\{(x_1 \lor b_1,1),(x_2 \lor b_2,1),(x_3,1),(x_4,1),(x_5,1),(x_6 \lor b_6,1),(\neg x_6 \lor b_7,1)\} \cup \phi_H$.

\item $mid=3$, the constraint $CNF(b_1+b_2+b_6+b_7 \leq 3)$ is included. $state=False$, $\phi_C \cap \phi_S=\{(x_3),(x_4)\}$, $B=\{b_1,b_2,b_3,b_4,b_6,b_7\}$, $\phi=\{(x_1 \lor b_1,1),(x_2 \lor b_2,1),(x_3 \lor b_3,1),(x_4 \lor b_4,1),(x_5,1),(x_6 \lor b_6,1),(\neg x_6 \lor b_7,1)\} \cup \phi_H$.

\item $mid=3$, the constraint $CNF(b_1+b_2+b_3+b_4+b_6+b_7 \leq 3)$ is included. $state=False$, $\phi_C \cap \phi_S=\{(x_1 \lor b_1),(x_2 \lor b_2),(x_3 \lor b_3),(x_4 \lor b_4),(x_6 \lor b_6),(\neg x_6 \lor b_7),(x_5)\}$, $B=\{b_1,b_2,b_3,b_4,b_5,b_6,b_7\}$, $\phi=\{(x_1 \lor b_1,1),(x_2 \lor b_2,1),(x_3 \lor b_3,1),(x_4 \lor b_4,1),(x_5 \lor b_5,1),(x_6 \lor b_6,1),(\neg x_6 \lor b_7,1)\} \cup \phi_H$.

\item $mid = 3$, $CNF(b_1+b_2+b_3+b_4+b_5+b_6+b_7 \leq 3)$ is included. $state=False$, $\phi_C \cap \phi_S=\{(x_1 \lor b_1),(x_2 \lor b_2),(x_3 \lor b_3),(x_4 \lor b_4),(x_5 \lor b_5),(x_6 \lor b_6),(\neg x_6 \lor b_7)\}$, $LB=3$.

\item $mid=5$, the constraint $CNF(b_1+b_2+b_3+b_4+b_6+b_7 \leq 5)$ is included. $state=True$, $I=\{x_1=False, x_2=False, x_3=True, x_4=False, x_5=True, x_6=False,b_1=True, b_2=True, b_3=False, b_4=True, b_5=False, b_6=True, b_7=False\}$, $UB=4$. The values of the $x_i,(1 \leq i \leq 6)$ variables in $I$ indeed constitute an optimal assignment.
\end{enumerate}
\end{exmp}

The core-guided binary search approach was improved by Heras\cite{heras2011core} \textit{et al.} with disjoint cores (see definition \ref{DisjointCore}).

\vspace{0.3in}
\begin{algorithm} [H]
\DontPrintSemicolon % Some LaTeX compilers require you to use \dontprintsemicolon instead
{\small
\KwIn{A WPMaxSAT instance $\phi = \phi_S \cup \phi_H$}
\KwOut{A WPMaxSAT solution to $\phi$}

\If{$SAT(\{C \mid (C,\infty)\in \phi_H \})=False$}{\Return{$\emptyset$}}

$\phi_W \gets \phi$\;
$\mathcal{C} \gets \emptyset$

\Repeat{$\forall_{C_i \in \mathcal{C}} UB_i \leq LB_i+1$}{
	\ForEach{$C_i \in \mathcal{C}$}{
		\If{$LB_i+1=UB_i$}{
			$mid_i \gets UB_i$\;
		}
		\Else{
			$mid_i \gets \lfloor \frac{LB_i+UB_i}{2} \rfloor$\;
		}
	}
		$(state,\phi_C,I)\gets SAT(\{C \mid (C,w)\in \phi_W \} \cup \bigcup_{C_i \in \mathcal{C}}CNF(\sum_{b_i \in B} w_ib_i \leq mid_i))$\;
		\If{$state=True$}{
			$lastI \gets I$\;
			\ForEach{$C_i \in \mathcal{C}$}{
				$UB_i \gets \sum_{b_r \in B} w_r(1-I(C_r \setminus \{b_r\})))$
			}
		}
		\Else{
			$subC \gets IntersectingCores(\phi_C,\mathcal{C})$\;
			\If{$\phi_C \cap \phi_S=\emptyset \text{ and }\vert sub\mathcal{C} \vert = 1$}{
				$LB \gets mid$ \tcp*[l]{$subC = \{(B,LB,mid,UB)\}$}
			}
			\Else{
			\ForEach{$(C_i,w_i) \in \phi_C \cap \phi_S$}{
				let $b_i$ be a new blocking variable\;
				$B \gets B \cup \{b_i\}$\;
				$\phi_W \gets \phi_W \setminus \{(C_i,w_i)\} \cup \{(C_i \lor b_i,w_i)\}$
			}
			$LB \gets 0$\;
			$UB \gets 1+\sum_{b_i \in B} w_i$\;
			\ForEach{$(B_i,LB_i,mid_i,UB_i) \in sub\mathcal{C}$}{
				$B \gets B \cup B_i$\;
				$LB \gets LB + LB_i$\;
				$UB \gets UB + UB_i$\;
			}
			$\mathcal{C} \gets \mathcal{C} \setminus sub\mathcal{C} \cup \{(B,LB,0,UB)\}$
			}
		}
}
\Return{$lastI$}
}
\caption{DisjointCoreGuided-BS$(\phi)$ Core-guided binary search extended with disjoint cores for solving WPMaxSAT.}
\label{algo:DisjointCoreGuidedBS}
\end{algorithm}
\vspace{0.3in}

Core-guided binary search methods with disjoint unsatisfiable cores maintains smaller lower and upper bounds for each disjoint core instead of just one global lower bound and one global upper bound. Thus, the algorithm will add multiple smaller cardinality constraints on the sum of the weights of the soft clauses rather than just one global constraint.

To maintain the smaller constraints, the algorithm keep information about the previous cores in a set called $\mathcal{C}$ initialized to $\emptyset$ (line 4) before the main loop. Whenever the SAT solver returns $False$ (line 12) it also provides a new core and a new entry $C_i=(B_i,LB_i,mid_i,UB_i)$ is added in $\mathcal{C}$ for $U_i$, where $B_i$ is the set of blocking variables associated with the soft clauses in $U_i$, $LB_i$ is a lower bound, $mid_i$ is the current middle value and $UB_i$ is an upper bound. The main loop terminates when for each $C_i \in \mathcal{C}$, $LB_i+1\geq UB_i$ (line 33). For each entry in $\mathcal{C}$, its middle value is calculated (lines 6-10) and a constraint for each entry is added to the working formula before calling the SAT solver on it (line 11). If the working formula is unsatisfiable (line 16), then, using $IntersectiongCores$, every core that intersects the current core is identified and its corresponding entry is added to $sub\mathcal{C}$ (line 17). If the core does not contain soft clauses that need to be relaxed and $\vert sub\mathcal{C} \vert=1$ (line 18), then $LB$ is assigned the value of the midpoint (line 19). On the other hand, if there exists clauses that has not been relaxed yet then the algorithm relaxes them (lines 21-24) and a new entry for the current core is added to $\mathcal{C}$ which accumulates the information of the previous cores in $sub\mathcal{C}$ (lines 25-31).

\begin{exmp}
Consider $\phi$ in example \ref{ex:LinearSearch} with all the weights of the soft clauses set to 1. At the beginning of algorithm \ref{algo:DisjointCoreGuidedBS}, we have $\phi_W = \phi$ and $\mathcal{C}=\emptyset$. The following are the iterations the algorithm executes.
\begin{enumerate}
\item No constraints to include. $state=False$, $\phi_C \cap \phi_S=\{(x_6), (\neg x_6)\}$, $sub\mathcal{C}=\emptyset$, $B=\{b_6,b_7\}$, $\phi_W=\{(x_1),(x_2),(x_3),(x_4),(x_5),(x_6\lor b_6),(\neg x_6 \lor b_7)\} \cup \phi_H$, $LB=0$, $UB=3$, $\mathcal{C}=\{(\{b_6,b_7\},0,0,3)\}$.

\item The constraint $CNF(b_6+b_7 \leq 1)$ is included. $state=False$, $\phi_C \cap \phi_S = \{(x_1),(x_2)\}$, $sub\mathcal{C}=\emptyset$, $B=\{b_1, b_2\}$. $\phi_W=\{(x_1 \lor b_1),(x_2 \lor b_2),(x_3),(x_4),(x_5), (x_6 \lor b_6), (\neg x_6 \lor b_7)\} \cup \phi_H$, $LB=0$, $UB=3$, $\mathcal{C}=\{(\{b_6,b_7\},0,0,3),(\{b_1,\\b_2\},0,0,3)\}$.

\item The constraints $\{CNF(b_6+b_7 \leq 1),CNF(b_1+b_2 \leq 1)\}$ are included. $state=False$, $\phi_C \cap \phi_S=\{(x_3), (x_4)\}$, $sub\mathcal{C}=\emptyset$, $B=\{b_3,b_4\}$, $\phi_W=\{(x_1 \lor b_1),(x_2 \lor b_2),(x_3 \lor b_3),(x_4 \lor b_4),(x_5), (x_6 \lor b_6), (\neg x_6 \lor b_7)\} \cup \phi_H$, $LB=0$, $UB=3$, $\mathcal{C}=\{(\{b_6,b_7\},0,0,3),\\(\{b_1,b_2\},0,0,3),(\{b_3,b_4\},0,0,3)\}$.

\item The constraints $\{CNF(b_6+b_7 \leq 1),CNF(b_1+b_2 \leq 1),CNF(b_3+b_4 \leq 1)\}$ are included. $state=False$, $\phi_C \cap \phi_S=\{(x_1 \lor b_1),(x_2 \lor b_2),(x_3 \lor b_3),(x_4 \lor b_4),(x_5)\}$, $sub\mathcal{C}=\{(\{b_1,b_2\},0,0,3),(\{b_3,b_4\},0,\\0,3)\}$, $B=\{b_1,b_2,b_3,b_4,b_5\}$, $\phi_W=\{(x_1 \lor b_1),(x_2 \lor b_2),(x_3 \lor b_3),(x_4 \lor b_4),(x_5 \lor b_5), (x_6 \lor b_6), (\neg x_6 \lor b_7)\} \cup \phi_H$, $LB=0$, $UB=8$, $\mathcal{C}=\{(\{b_6,b_7\},0,0,3),(\{b_1,b_2,b_3,b_4,b_5\},0,0,8)\}$.

\item The constraints $CNF(b_6+b_7\leq 1),CNF(b_1+b_2+b_3+b_4+b_5\leq 4)$ are included. $state=True$, $I=\{x_1=False,x_2=False,x_3=True,x_4=False,x_5=True,x_6=False,b_1=True,b_2=True,b_3=False,b_3=True,b_5=False,b_6=True,b_7=False\}$, $\mathcal{C}=\{(\{b_6,b_7\},0,0,1),(\{b_1,b_2,b_3,b_4,b_5\},0,0,2)\}$.

\item The constraints $CNF(b_6+b_7\leq 1),CNF(b_1+b_2+b_3+b_4+b_5\leq 1)$ are included. $state=False$, $\phi_C \cap \phi_S=\{(x_1 \lor b_1),(x_2 \lor b_2),(x_3 \lor b_3),(x_4 \lor b_4),(x_5 \lor b_5)\}$, $sub\mathcal{C}=\{ (\{b_1,b_2,b_3,b_4,b_5\},0,0,2) \}$, $\mathcal{C}=\{(\{b_6,b_7\},0,0,1),(\{b_1,b_2,b_3,b_4,b_5\},1,0,2)\}$.

\item $state=True$, $I=\{x_1=False,x_2=False,x_3=True,x_4=False,x_5=True,x_6=False,b_1=True,b_2=True,b_3=False,b_4=True,b_5=False,b_6=True,b_7=False\}$.
\end{enumerate}
\end{exmp}

SAT-based WPMaxSAT solvers rely heavily on the hardness of the SAT formulae returned by the underlying SAT solver used. Obviously, the location of the optimum solution depends on the structure of the instances returned and the number of iterations it takes to switch from $True$ to $False$ (or from $False$ to $True$).

\section{Portfolio MaxSAT Techniques}
The results of the MaxSAT Evaluations suggest there is no absolute best algorithm for solving MaxSAT. This is because the most efficient solver often depends on the type of instance. In other words, different solution approaches work well on different families of instances\cite{matos2008max}. Having an oracle able to predict the most suitable MaxSAT solver for a given instance would result in the most robust solver. The success of SATzilla\cite{xu2008satzilla} for SAT was due to a regression function which was trained to predict the performance of every solver in the given set of solvers based on the features of an instance. When faced with a new instance, the solver with the best predicted runtime is run on the given instance. The resulting SAT portfolios excelled in the SAT Competitions in 2007 and in 2009 and pushed the state-of-the-art in SAT solving. When this approach is extended to (WP)MaxSAT, the resulting portfolio can achieve significant performance improvements on a representative set of instances.

ISAC\cite{ansotegui2014maxsat} (Instance-Specific Algorithm Configuration) is one of the most successful WPMaxSAT portfolio algorithms. It works by computing a representative feature vector that characterizes the given input instance in order to identify clusters of similar instances. The data is therefore clustered into non-overlapping groups and a single solver is selected for each group based on some performance characteristic. Given a new instance, its features are computed and it is assigned to the nearest cluster. The instance is then solved by the solver assigned to that cluster.

\section{Translating Pseudo-Boolean Constraints into CNF}
This section discusses translating pseudo-Boolean (PB) constraints into CNF. The procedure is needed in almost every SAT-based WPMaxSAT algorithm and its efficiency surely affects the overall performance of the solver.

\subsection{Introduction}
A \textit{PB constraint} is a linear constraint over Boolean variables. PB constraints are intensively used in expressing NP-hard problems. While there are dedicated solvers (such as Sat4j) for solving PB constraints, there are good reasons to be interested in transforming the constraints into SAT (CNF formulae), and a number of methods for doing this have been reported\cite{sinz2005towards,bailleux2003efficient,marques2007towards,aavani2013new,steinke2014pblib,manthey2014more,aavani2011translating,bailleux2006translation}.

\begin{defn}[PB constraint]
A PB constraint is an inequality (equality) on a linear combination of Boolean literals $l_i$
$$\sum_{i=1}^n a_i l_i \{<,\geq,=,\leq,> \} K$$
where $a_1,\dots,a_n$ and $K$ (called the bound) are constant integers and $l_1,\dots,l_n$ are literals.
\end{defn}

There are at least two clear benefits of solving PB constraints by encoding them into CNF. First, high-performance SAT solvers are being enhanced continuously, and since they take a standard input format  there is always a selection of good solvers to make use of. Second, solving problems involving Boolean combinations of constraints is straightforward. This approach is particularly attractive for problems which
are naturally represented by a relatively small number of PB
constraints (like the Knapsack problem) together which a large number of purely Boolean constraints.

\subsection{Encoding method}
We present the method of Bailleux, Boufkhad and Roussel\cite{bailleux2006translation}. In their paper, they consider (without loss of generality) PB constraints of the form $\sum_{i=1}^n a_il_i \leq K$, where $a_1 \leq a_2 \leq \dots \leq a_n$. This type of constraint is denoted by the triple $\langle A_n, L_n, K \rangle$, where $A_n=(a_1,\dots,a_n)$ and $L_n=(l_1,\dots,l_n)$. For some bound $b$, the triple $\langle A_i,L_i,b \rangle$, for $1 \leq i \leq n$, represents the PB constraint $a_il_i + a_2l_2 + \dots + a_il_i \leq b$. When the tuples $A_n$ and $L_n$ are fixed, a triple $\langle A_i,L_i,b \rangle$ representing a PB constraint is defined with no ambiguity by the integer $i$ and the bound $b$.

For each $\langle A_i, L_i, b \rangle$, a new variable $D_{i,b}$ is introduced. This new variable represents the satisfaction of the constraint $\langle A_i, L_i, b \rangle$, i.e., $D_{i,b}=True$ if and only if $\langle A_i, L_i, b \rangle$ is satisfied. The variable $D_{n,K}$ represents $\langle A_n, L_n, K \rangle$ and the correctness of the encoding is conditioned by the fact that an assignment satisfies $\langle A_n, L_n, K \rangle$ if and only if it satisfies the encoded CNF formula and fixes $D_{n,K}$ to $True$.

The variables $D_{i,b}$ such that $b \leq 0$ or $b \geq \sum_{j=1}^i a_j$ are called \textit{terminal variables}.

The encoding starts with a set of variables containing the original variables PB constraint and the variable $D_{n,K}$. The variables $l_i$ are marked. At each step, an unmarked variable $D_{i,b}$ is considered. If $D_{i,b}$ is not terminal the two variables $D_{i-1,b}$ and $D_{i-1,b-a_i}$ are added to the set of variables if they are not already in it and the following four clauses are added
$$(\neg D_{i-1,b-a_i} \lor D_{i,b}),(D_{i,b} \lor D_{i-1,b}), (D_{i,b} \lor l_i \lor D_{i-1,b-a_i}), (D_{i-1,b} \lor l_i \lor D_{i,b})$$

Next, $D_{i,b}$ is marked so it won't be considered again.

In case that $D_{i,b}$ is a terminal variable, then by definition either $b \leq 0$ or $b \geq \sum_{j=1}^i a_j$ and $D_{i,b}$ is fixed as follows
\[D_{i,b} = 
\begin{cases} 
      False & \text {if } b < 0. \text{ The clause }\neg D_{i,b} \text{ is added to the formula.} \\
      True & \text{if }\sum_{j=1}^i a_j \leq b. \text{ The clause } D_{i,b} \text{ is added to the formula.}
   \end{cases}
\]
When $b = False$, every variable in the constraint must be set to $False$. To achieve this, for every $1 \leq j \leq i$, the clauses $(D_{i,0} \lor l_j)$ are added together with the clause $(l_1 \lor l_2 \lor \dots \lor D_{i,0})$. The procedure stops when there are no more unmarked variables.

\begin{exmp}
This example illustrates the encoding of the PB constraint $2x_1 + 3x_2 + 4x_3 \leq 6$. The formula $\phi = \{(\neg D_{2,2} \lor D_{3,6}),(\neg D_{3,6} \lor \neg x_3 \lor D_{2,2}),(\neg D_{2,6} \lor x_3 \lor D_{3,6}),(D_{2,6} \lor \neg D_{1,-1} \lor D_{2,2}),(\neg D_{2,2} \lor D_{1,2}),(\neg D_{2,2} \lor \neg x_2 \lor D_{1,-1}),(\neg D_{1,2} \lor x_2 \lor D_{2,2}),(D_{1,2}),(\neg D_{1,-1})\}$. Thus, $D_{3,6}=True$ only if at least one of $x_2$ or $x_3$ is $False$.
\end{exmp}

The correctness and the complexity of the encoding are discussed in the same paper\cite{bailleux2006translation}.

\subsection{Complexity of the encoding}
The complexity of the encoding is measured in terms of the number of variables. The number of clauses produced is related by a constant factor to the number of variables. There are cases where the previous procedure produces a polynomial and others that produce an exponential number of variables.

\subsubsection{Polynomial cases}
\begin{figure}[h!]
\centering
\parbox{5cm}{
\centering
\includegraphics[scale=0.6]{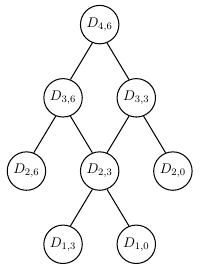}
\caption{Variables introduced to encode $3x_1+3x_2+3x_3+3x_4 \leq 6$.}
\label{Fig:card1}}
\qquad
\begin{minipage}{5cm}
\centering
\includegraphics[scale=0.57]{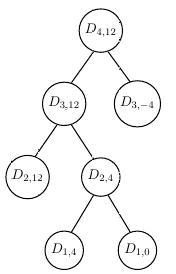}
\caption{Variables introduced to encode $2x_1+ 4x_2+ 8x_3+ 16x_4 \leq 12$.}
\label{Fig:card2}
\end{minipage}
\end{figure}

The encoding seems to generate an exponential number of variables: at each step a non-terminal variable creates two variables that will in turn create two other variables each and so on. However, this is not true for terminal variables and for variables that have already been considered by the procedure. When a terminal variable is met, it is said to be a \textit{cut} in the procedure and when a variable already in the set of variables is met, it is said to have \textit{merged} in the procedure. By the cuts and merges, the size of encodings can be polynomial in some cases. There are two restrictions on the PB constraint for it to have a polynomial-size encoding:
\begin{enumerate}
\item The integers $a_i$'s are bounded by a polynomial in $n$, $P(n)$. In this case, the potential number of $D_{i,b}$ variables for some $i$ is $2^{n-i}$ but because of the merges, this number reduces to a polynomial since the variables $D_{i,b}$ for some $i$ are such that $m \leq b \leq M$ where $m$ is at least equal to $K - \sum_{j=0}^i a_{n-j}$ and $M \leq K$, $b$ can take at most $M ? m$ different values and then it can take at most $\sum_{j=0}^i a_{n?j}$ different values, which is bounded by $(n-i)P(n)$. Since there are $n$ different possible values for $i$, the total number of variables is bounded by a polynomial in $n$. Figure \ref{Fig:card1} shows an example of this case.

\item The weights are $a_i= \alpha_i$ where $\alpha \geq 2$. In this case, for every non terminal variable $D_{i,b}$ considered in the procedure, at least one of the variables $D_{i-1,b}$ or $D_{i-1,b?\alpha_i}$ is a terminal variable. This is true because $\sum_{j=0}^{i?1} \alpha^j < \alpha_i$. Either $b \geq \alpha^i$ and then $\sum_{j=0}^{i-1} \alpha^j < b$ and then $D_{i-1,b} $ is a terminal variable or $b < \alpha^i$ and in this case $D_{i-1,b-\alpha^i}$ is a terminal variable. Thus, there is a cut each time a variable is considered in the procedure. Figure \ref{Fig:card2} shows an example for this case.
\end{enumerate}

\begin{figure}[h!]
\centering
\includegraphics[scale=0.5]{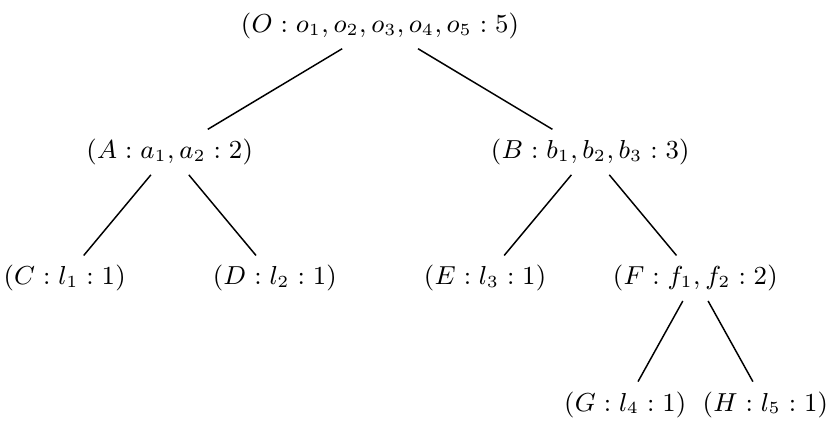}
\caption{Totalizer encoding for the constraint $l_1 + \dots + l_5 \leq k$}
\label{Fig:totalizer}
\end{figure}

\subsubsection{Exponential cases}
There are possible sequences of $a_i$'s that will give a tree with branches of length $\Omega(n)$ and with no possible merge of nodes (which implies a tree of size $\Omega(2n)$). The idea here is simply to combine a constant sequence with a geometric sequence. Let $n$ be the length of the PB constraint $Q$ and let $a_i = \alpha + b^i$ such that $\alpha = b^{n+2}$. The key point is that the geometric term must be negligible compared to the constant term, that is $\sum_{i=0}^n b^i < \alpha$. For simplicity, we will choose $b = 2$. Note that in this case, $a_i= 2^{n+2} + 2^i$ which is not bounded by a polynomial in $n$. Fix $K = \alpha \times \frac{n}{2} = n\times 2^{n+1}$.

A terminal node is reached when we get a term $D_{i,k}$ such that $k \leq 0$ or $k \geq \sum_{j=1}^i a_j$. Because the constant term is predominant, the first condition cannot be met before $i = \frac{K}{\alpha} = \frac{n}{2}$. The earliest case where the second condition can be satisfied is when $k$ remains equal to $K$. We have $\sum_{j=1}^i a_j = \sum_{j=1}^i \alpha+b^j = \alpha \times i + \sum_{j=1}^i b^j \geq \alpha \times i$. Therefore, the earliest case where the second condition can be met is when $\alpha \times \frac{n}{2} = \alpha \times i$ which means $i = \frac{n}{2}$. We can conclude that each branch is at least of length $\frac{n}{2}$.

In addition, in the encoding, each node of the tree holds the term $D_{i,k}$ which corresponds to $\sum_{j=1}^i a_jx_j \leq K - \sum_{j\in S} a_j$, where $S \subset [i+1..n]$. One key point is that in the binary representation of $K-\sum_{j\in S} a_j$, the $n$ least significant bits directly correspond to the indices in $S$. Therefore, these $n$ least significant bits of the right term are necessarily different from one node to another. For this reason, no node can be merged. Because of this and since branches are of length at least equal to $\frac{n}{2}$, the size of the tree is at least $2^{\frac{n}{2}}$ and the encoding of this particular constraint is of exponential size.

\subsection{Other encoding techniques}
Incremental approaches\cite{morgado2014core,martins2014incremental,narodytska2014maximum} allow the constraint solver to retain knowledge from previous iterations that may be used in the upcoming iterations. The goal is to retain the inner state of the constraint solver as well as learned clauses that were discovered during the solving process of previous iterations. At each iteration, most MaxSAT algorithms create a new instance of the constraint solver and rebuild the formula losing most if not all the knowledge that could be derived from previous iterations.

\section{Experimental Investigation}
We conducted an experimental investigation in order to compare the performance of different WPMaxSAT solvers to branch and bound solvers on a number of benchmarks  instances.

Experimental evaluations of MaxSAT solvers has gained great interest among SAT and MaxSAT researchers. This is due to the fact that solvers are becoming more and more efficient and adequate to handle WPMaxSAT instances coming from real-life applications. Thus, carrying out such an investigation and comparing the efficiency of different solvers is critical to knowing which solving technique is suitable for which category of inputs. In fact, an annual event called the \href{http://www.maxsat.udl.cat/}{MaxSAT Evaluations} is scheduled just for this purpose. The \href{http://www2.iiia.csic.es/conferences/maxsat06/}{first MaxSAT Evaluation} was held in 2006. The objective of the MaxSAT Evaluation is comparing the performance of state of the art (weighted) (partial) MaxSAT solvers on a number of benchmarks and declaring a winner for each benchmark category.

The solvers that we investigate participated in the MaxSAT Evaluations of \href{http://www.maxsat.udl.cat/13/introduction/index.html}{2013} and \href{http://www.maxsat.udl.cat/14/index.html}{2014}. A number of the solvers are available online while some of them were not and we had to contact the authors to get a copy. The benchmarks we used participated in the 2013 MaxSAT Evaluation and are WPMaxSAT instances of three categories: random, crafted and industrial.

The solvers were run on a machine with an Intel\circledR\,Core\texttrademark\, i5 CPU clocked at 2.4GHz, with 5.7GB of RAM running elementary OS Linux. The timeout is set to 1000 seconds and running the solvers on the benchmarks took roughly three months. We picked elementaryOS because it does not consume too many resources to run and thus giving enough room for the solvers to run. In addition, elementaryOS is compatible with popular Ubuntu distribution which makes it compatible with its repositories and packages.

\subsection{Solvers descriptions}
\label{section:solverdesc}
The solvers we experimented with are:
\begin{enumerate}
\item \textbf{WMiFuMax} is an unsatisfiability-based WPMaxSAT solver based on the technique of Fu and Malik\cite{fu2006solving} and on the algorithm by Manquinho, Marques-Silva, and Planes\cite{manquinho2009algorithms}, which is works by identifying unsatisfiable sub-formulae. MiFuMax placed third in the WPMaxSAT industrial category of the 2013 MaxSAT evaluation. The solver (and the source code) is available online under the GNU General Public License. The SAT solver used is called MiniSAT\cite{sorensson2005minisat}. Author: Mikol\'{a}\v{s} Janota.

\item \textbf{QWMaxSAT} is a weighted version of QMaxSAT developed by Koshimura, Zhang, Fujita and Hasegawa\cite{koshimura2012qmaxsat} and is available freely online. This solver is a satisfiability-based solver built on top of version 2.0 of the SAT solver MiniSAT\cite{een2005minisat}. The authors of QMaxSAT modified only the top-level part of MiniSat to manipulate cardinality constraints, and the other parts remain unchanged. Despite originally being a PMaxSAT solver, the authors developed a version of the solver for WPMaxSAT in 2014. Authors: Miyuki Koshimura, Miyuki Koshimura, Hiroshi Fujita and Ryuzo Hasegawa.

\item \textbf{Sat4j}\cite{le2010sat4j} is a satisfiability-based WPMaxSAT solver developed by Le Berre and Parrain. The solver works by translating WPMaxSAT instances into pseudo-Boolean optimization ones. The idea is to add a blocking variable per weighted soft clause that represents that such clause has been violated, and to translate the maximization problem on those weighted soft clauses into a minimization problem on a linear function over those variables. Given a WPMaxSAT instance $\phi= \{(C_1,w_1),\dots,(C_n,w_n)\} \cup \phi_H$, Sat4j translates $\phi$ into $\phi'=\{(C_1\lor b_1),\dots,(C_n\lor b_n) \}$ plus an objective function $min:\sum_{i=1}^n w_ib_i$. Sat4j avoids adding blocking variables to both hard and unit clauses. the Sat4j framework includes the pseudo-Boolean solver Sat4j-PB-Res which is used to solve the encoded WPMaxSAT problem. Authors: Daniel Le Berre and Emmanuel Lonca.

\item \textbf{MSUnCore}\cite{marques2009msuncore} is an unsatisfiability-based WPMaxSAT solver built on top the SAT solver PicoSAT\cite{biere2008picosat}. This solver implements a number of algorithms capable of solving MaxSAT, PMaxSAT and W(P)MaxSAT. MSUnCore uses PicoSAT for iterative identification of unsatisfiable cores with larger weights. Although ideally a minimal core would be preferred, any unsatisfiable core can be considered. Clauses in identified core are then relaxed by adding a relaxation variable to each clause. Cardinality constraints are encoded using several encodings, such as the pairwise and bitwise encodings\cite{prestwich2009cnf,prestwich2007variable}, the ladder encoding\cite{gent2004new}, sequential counters\cite{sinz2005towards}, sorting networks\cite{een2006translating}, and binary decision diagrams (BDDs)\cite{een2006translating}. Authors: Ant\'{o}nio Morgado, Joao Marques-Silva, and Federico Heras.

\item \textbf{Maxsatz2013f} is a very successful branch and bound solver that placed first in the WPMaxSAT random category of the 2013 MaxSAT evaluation. It is based on an earlier solver called Maxsatz\cite{li2009exploiting}, which incorporates the technique developed for the famous SAT solver, Satz\cite{li1997heuristics}. At each node, it transforms the instance into an equivalent one by applying efficient refinements of unit resolution ($(A \lor B)$ and $(\neg B)$ yield $A$) which replaces $\{(x), (y), (\neg x \lor \neg y)\}$ with $\{\Box, (x \lor y)\}$ and $\{(x), (\neg x \lor y), (\neg x \lor z), (\neg y \lor \neg z)\}$ with $\{ \Box, (\neg x \lor y \lor z), (x \lor \neg y \lor \neg z)\}$. Also, it implements a lower bound method (enhanced with failed literal detection) that increments the lower bound by one for every disjoint inconsistent subset that is detected by unit propagation. The variable selection heuristics takes into account the number of positive and negative occurrences in binary and ternary clauses. Maxsatz2013f is available freely online. Authors: Chu Min Li, Yanli LIU, Felip Many\`{a}, Zhu Zhu and Kun He.

\item \textbf{WMaxSatz-2009} and \textbf{WMaxSatz+}\cite{li2007new,li2010resolution} are branch and bound solvers that use transformation rules\cite{li2009exploiting} which can be implemented efficiently as a by-product of unit propagation or failed literal detection. This means that the transformation rules can be applied at each node of the search tree. Authors: Josep Argelich, Chu Min Li, Jordi Planes and Felip Many\`{a}.

\item \textbf{ISAC+}\cite{ansotegui2014maxsat} (Instance-Specific Algorithm Configuration) is a portfolio of algorithm which, given a WPMaxSAT instance, selects the solver better suited for that instance. A regression function is trained to predict the performance of every solver in the given set of solvers based on the features of an instance. When faced with a new instance, the solver with the best predicted runtime is run on the given instance. ISAC+ uses a number of branch and bound solvers as well as SAT-based, including QMaxSAT, WMaxSatz-2009 and WMaxSatz+. Authors: Carlos Ans\'{o}tegui, Joel Gabas, Yuri Malitsky and Meinolf Sellmann.
\end{enumerate}

\begin{center}
\begin{tabular}{ |l|l|l| }
\hline
\multicolumn{3}{ |c| }{\textbf{Summary}} \\
\hline
\textbf{Technique} & \textbf{Solver name} & \textbf{Sub-technique} \\ \hline
\multirow{4}{*}{Satisfiability-based} & WMiFuMax & SAT-based \\
 & QWMaxSAT & SAT-based \\
 & Sat4j & SAT-based \\
 & MSUnCore & UNSAT-based \\ \hline
\multirow{3}{*}{Branch and bound} & Maxsatz2013f & \\
 & WMaxSatz-2009 & \\
 & WMaxSatz+ & \\ \hline
Portfolio & ISAC+ & \\ \hline

\hline
\end{tabular}
\end{center}

\subsection{Benchmarks descriptions}
\label{section:benchmarkdesc}
The benchmarks we used are the \href{http://www.maxsat.udl.cat/13/benchmarks/index.html}{WPMaxSAT instances} of the 2013 MaxSAT Evaluation and are divided into three categories:
\begin{enumerate}
\item Random: This category consists of WPMax-2-SAT and WPMax-3-SAT instances generated uniformly at random. The WPMax-2-SAT instances are divided into formulae with low (lo), medium (me) and high (hi) numbers of variables and clauses. The WPMax-3-SAT instances contain three literals per clause and have a high number of variables and clauses.

\item Crafted: These instances are specifically designed to give a hard time to the solver. There is an award for the smallest instance that can not be solved by any solver.

\item Industrial: Consists of instances that come from various applications of practical interest, such as model checking, planning, encryption, bio-informatics, etc. encoded into MaxSAT. This category is intended to provide a snapshot of the current strength of solvers as engines for SAT-based applications.
\end{enumerate} 
In the MaxSAT Evaluations, a first, second and third place winners are declared for each of the three categories.

\subsection{Results}
\label{section:results}
In this section, the results we obtained are presented and discussed. For each category, we present the constituting sets of instances and their sizes, the number of instances solved by each solver and the amount of time it took each solver to work on each set of instances.
\subsubsection{Random category}
The three sets of instances in the random category are:
\begin{table}[H]
\centering
\begin{tabular}{|l|c|c|}
\hline
\textbf{Name} & \multicolumn{1}{l|}{\textbf{Abbreviation}} & \multicolumn{1}{l|}{\textbf{\# of instances}} \\ \hline
wpmax2sat-lo  & lo                                         & 30                                            \\ \hline
wpmax2sat-me  & me                                         & 30                                            \\ \hline
wpmax2sat-hi  & hi                                         & 30                                            \\ \hline
wpmax3sat-hi  & 3hi                                        & 30                                            \\ \hline
\end{tabular}
\end{table}

\begin{table}[H]
\centering
\begin{tabular}{|l|c|c|c|c|}
\hline
\textbf{Solver}    & \multicolumn{1}{l|}{\textbf{lo}} & \multicolumn{1}{l|}{\textbf{me}} & \multicolumn{1}{l|}{\textbf{hi}} & \multicolumn{1}{l|}{\textbf{3hi}} \\\hline
{MiFuMax}          & 0                 & 0            & 0            & 0            \\\hline
{QWMaxSAT}         & 0                 & 0            & 0            & 0            \\\hline
{Sat4j}            & 0                 & 0            & 0            & 0            \\\hline
{MSUnCore}         & 0                 & 0            & 0            & 0            \\\hline
{MaxSatz2013f}     & 30                & 30           & 29           & 30           \\\hline
{WMaxSatz-2009}    & 30                & 30           & 29           & 30           \\\hline
{WMaxSatz+}        & 30                & 30           & 29           & 30           \\\hline
{ISAC+}            & 29                & 8            & 1            & 10          \\ \hline
\end{tabular}
\caption{Number of instances solved in the random category.}
\label{table:random}
\end{table}

\begin{table}[H]
\centering
\begin{tabular}{|l|c|c|c|c|c|}
\hline
\textbf{Solver}  & \textbf{lo} & \textbf{me} & \textbf{hi} & \textbf{3hi} & \textbf{Total} \\ \hline
WMiFuMax         & 0\%         & 0\%         & 0\%         & 0\%          & 0\%            \\ \hline
QWMaxSAT         & 0\%         & 0\%         & 0\%         & 0\%          & 0\%            \\ \hline
Sat4j            & 0\%         & 0\%         & 0\%         & 0\%          & 0\%            \\ \hline
MSUnCore         & 0\%         & 0\%         & 0\%         & 0\%          & 0\%            \\ \hline
MaxSatz2013f     & 100\%       & 100\%       & 96.7\%      & 100\%        & 99.2\%         \\ \hline
WMaxSatz-2009    & 100\%       & 100\%       & 96.7\%      & 100\%        & 99.2\%         \\ \hline
WMaxSatz+        & 100\%       & 100\%       & 96.7\%      & 100\%        & 99.2\%         \\ \hline
ISAC+            & 96.7\%      & 26.7\%      & 3.3\%       & 33.3\%       & 40\%           \\ \hline
\end{tabular}
\caption{Percentages of instances solved in the random category.}
\label{table:rndpercentage}
\end{table}

\begin{figure}[H]
\centering
\includegraphics[scale=0.52]{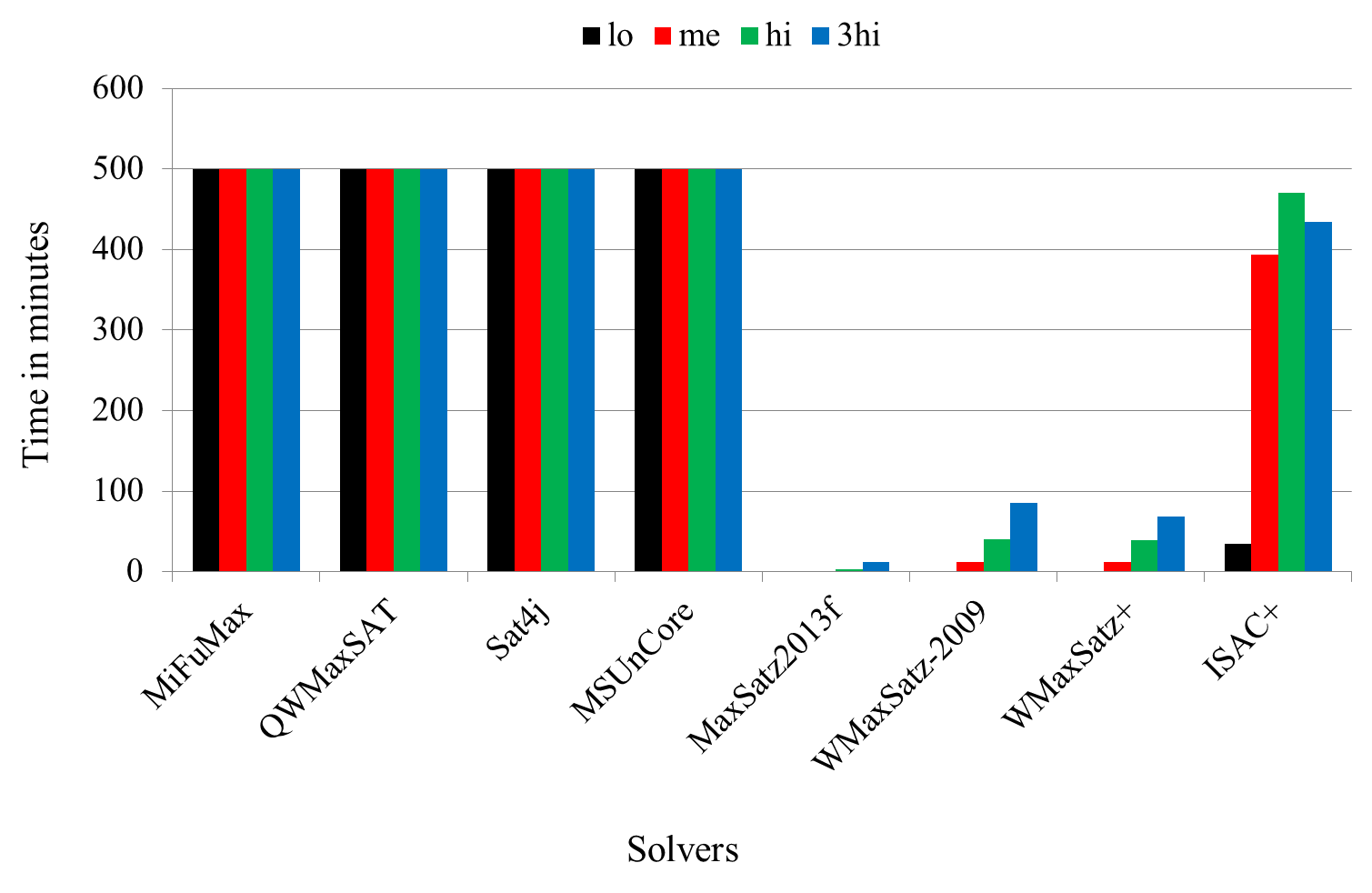}
\caption{Time results for the random category.}
\label{fig:random1}
\end{figure}

The branch and bound solvers MaxSatz2013f, WMaxSatz-2009 and WMaxSatz+ performed considerably better than the SAT-based solvers in the random category. In particular, MaxSatz2013f finished the four benchmarks under 16 minutes, while WMiFuMax, MSUnCore and Sat4j timedout on most instances. MaxSatz2013f placed first in the random category in the 2013 MaxSAT Evaluation, see {\footnotesize\url{http://www.maxsat.udl.cat/13/results/index.html#wpms-random-pc}}. The top non branch and bound solver is ISAC+, which placed third in the random category in 2014 (see {\footnotesize \url{http://www.maxsat.udl.cat/14/results/index.html#wpms-random-pc}}).

\subsubsection{Crafted category}
The seven sets of instances in the crafted category are:
\begin{table}[H]
\centering
\begin{tabular}{|l|c|c|}
\hline
\textbf{Name} & \multicolumn{1}{l|}{\textbf{Abbreviation}} & \multicolumn{1}{l|}{\textbf{\# of instances}} \\ \hline
auctions/auc-paths  & auc/paths     & 86                                            \\ \hline
auctions/auc-scheduling & auc/sch & 84 \\ \hline
CSG  & csg          & 10                                            \\ \hline
min-enc/planning  & planning      & 56                                            \\ \hline
min-enc/warehouses & warehouses       & 18                                            \\ \hline
pseudo/miplib	& miplib & 12 \\ \hline
random-net & rnd-net & 74 \\ \hline
\end{tabular}

\label{table:crafteddesc}
\end{table}

\begin{table}[H]
{\scriptsize

\begin{center}
\begin{tabular}{|l|c|c|c|c|c|c|c|}

\hline
\textbf{Solver} & \textbf{auc/paths} & \textbf{auc/sch} &\textbf{csg} & \textbf{planning} & \textbf{warehouses} & \textbf{miplib} & \textbf{rnd-net} \\ \hline
WMiFuMax         & 84      & 84    & 5            & 23                & 0                   & 1               & 8                   \\ \hline
QWMaxSAT        & 84      &  84   & 10           & 56                & 2                   & 4               & 1                   \\ \hline
Sat4j           & 55     &   55   & 10           & 56                & 1                   & 4               & 0                   \\ \hline
MSUnCore        & 84      &  84   & 6            & 53                & 0                   & 0               & 0                   \\ \hline
MaxSatz2013f    & 81     &   81   & 1            & 41                & 6                   & 4               & 1                   \\ \hline
WMaxSatz-2009   & 67     &   67   & 1            & 45                & 6                   & 3               & 0                   \\ \hline
WMaxSatz+       & 66     &   66   & 1            & 45                & 6                   & 2               & 0                   \\ \hline
ISAC+           & 84      &  84   & 4            & 53                & 18                  & 3               & 55                  \\ \hline
\end{tabular}
\end{center}
}
\caption{Number of instances solved by each solver.}
\label{table:crafted}

\end{table}

\begin{table}[H]
{\scriptsize
{\fontsize{7}{1}

\centering

\begin{tabular}{|l|c|c|c|c|c|c|c|c|}

\hline

\textbf{Solver} & \textbf{auc/paths} & \textbf{auc/sch} & \textbf{csg} & \textbf{planning} & \textbf{warehouses} & \textbf{miplib} & \textbf{rnd-net} & \textbf{Total} \\ \hline
WMiFuMax        & 2.3\%              & 100\%            & 50\%         & 41.1\%            & 0\%                 & 8.3\%           & 10.8\%   & 30.1\%           \\ \hline
QWMaxSAT        & 52.3\%             & 100\%            & 100\%        & 100\%             & 11.1\%              & 33.3\%          & 1.4\%   & 57\%            \\ \hline
Sat4j           & 31.4\%             & 65.5\%           & 100\%        & 100\%             & 5.6\%               & 33.3\%          & 0\%     & 48\%           \\ \hline
MSUnCore        & 16.3\%             & 100\%            & 60\%         & 94.6\%            & 0\%                 & 0\%             & 0\%    & 38.7\%              \\ \hline
MaxSatz2013f    & 100\%              & 96.4\%           & 10\%         & 73.2\%            & 33.3\%              & 33.3\%          & 1.4\%      & 49.7\%         \\ \hline
WMaxSatz-2009   & 100\%              & 79.8\%           & 10\%         & 80.4\%            & 33.3\%              & 25\%            & 0\%    & 47\%             \\ \hline
WMaxSatz+       & 100\%              & 78.6\%           & 10\%         & 80.4\%            & 33.3\%              & 16.7\%          & 0\%      & 45.6\%            \\ \hline
ISAC+           & 100\%              & 100\%            & 40\%         & 94.6\%            & 100\%               & 25\%            & 74.3\%   & 76.3\%           

\\ \hline

\end{tabular}
}}
\caption{Percentages of instances solved in the crafted category.}
\end{table}

\begin{figure}[H]
\centering
\includegraphics[scale=0.53]{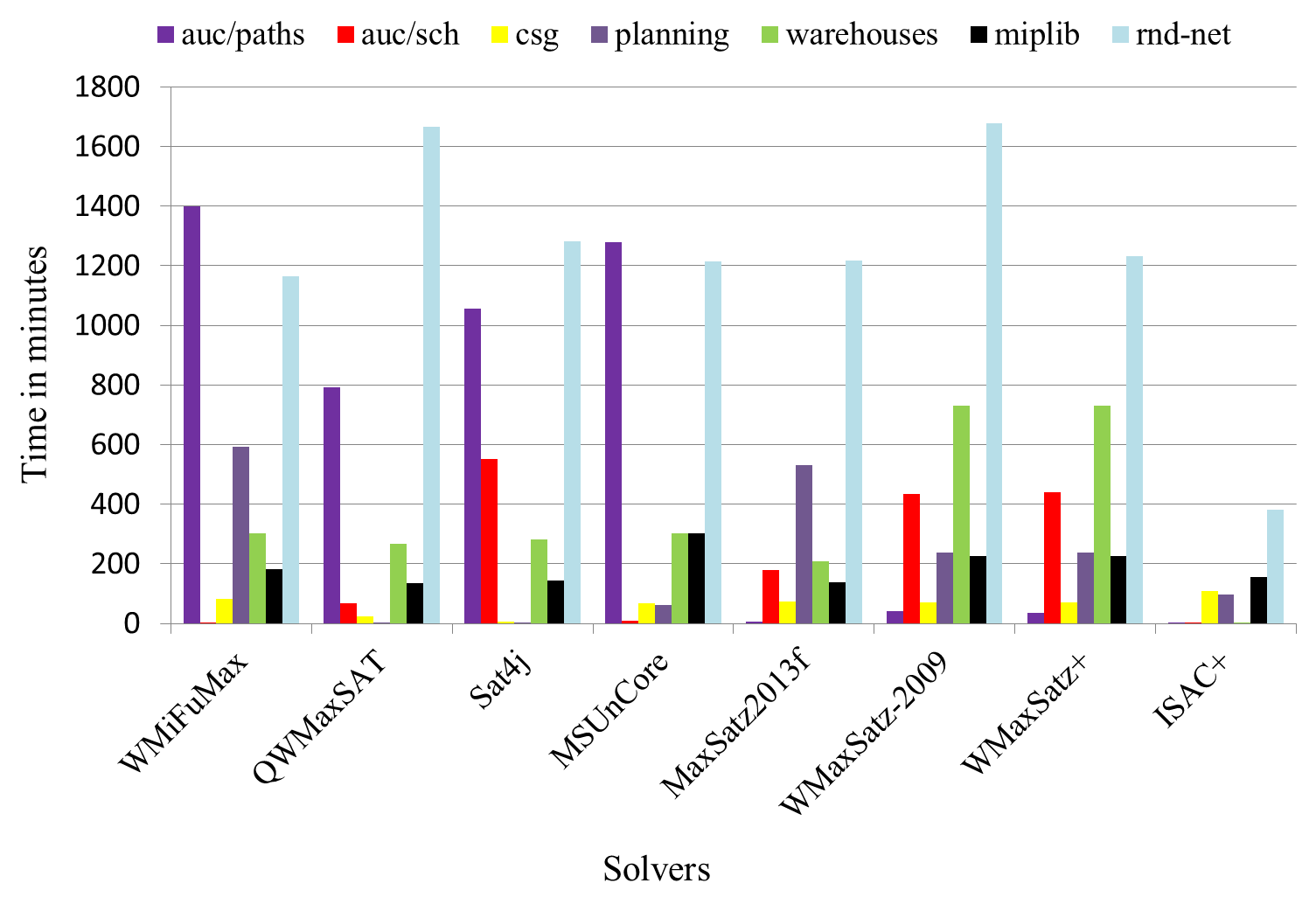}
\caption{Time results for the crafted category.}
\label{fig:crafted1}
\end{figure}

As it can be noticed from the results, ISAC+ is the winner of the crafted category. Indeed, the winner of this category in the 2014 MaxSAT Evaluation is ISAC+ (see {\footnotesize \url{http://www.maxsat.udl.cat/14/results/index.html#wpms-crafted}}), and in the 2013 evaluation it placed second (see {\footnotesize \url{http://www.maxsat.udl.cat/13/results/index.html#wpms-crafted-pc}}). Generally, SAT-based and branch and bound solvers perform nearly equally on crafted instances.

\subsubsection{Industrial category}
The seven sets of instance in the industrial category are:
\begin{table}[H]
\centering
\begin{tabular}{|l|c|c|}
\hline
\textbf{Name}          & \textbf{Abbreviation} & \textbf{\# of instances} \\ \hline
wcsp/spot5/dir         & wcsp-dir              & 21                       \\ \hline
wcsp/spot5/log         & wcsp-log              & 21                       \\ \hline
haplotyping-pedigrees  & HT                    & 100                      \\ \hline
upgradeability-problem & UP                    & 100                      \\ \hline
preference\_planning   & PP                    & 29                       \\ \hline
packup-wpms            & PWPMS                 & 99                       \\ \hline
timetabling            & TT                    & 26                       \\ \hline
\end{tabular}
\end{table}

 \begin{table}[H]
 {\footnotesize
 \begin{center}
 \begin{tabular}{|l|c|c|c|c|c|c|c|}
 \hline
 \textbf{Solver} & \textbf{wcsp-dir} & \textbf{wcsp-log} & \textbf{HT} & \textbf{UP} & \textbf{PP} & \textbf{PWPMS} & \textbf{TT} \\ \hline
 WMiFuMax         & 6                 & 6                 & 85          & 100         & 11          & 46             & 0           \\ \hline
 QWMaxSAT        & 14                & 13                & 20          & 0           & 29          & 17             & 8           \\ \hline
 Sat4j           & 3                 & 3                 & 15          & 37          & 28          & 2              & 8           \\ \hline
 MSUnCore        & 14                & 14                & 89          & 100         & 25          & 0              & 0           \\ \hline
 MaxSatz2013f    & 4                 & 4                 & 0           & 0           & 5           & 25             & 0           \\ \hline
 WMaxSatz-2009   & 4                 & 3                 & 0           & 41          & 5           & 12             & 0           \\ \hline
 WMaxSatz+       & 4                 & 3                 & 0           & 41          & 5           & 12             & 0           \\ \hline
 ISAC+           & 17                & 7                 & 15          & 100          & 9           & 99            & 9           \\ \hline
 \end{tabular}
 \end{center}
 }
 \caption{Number of instances solved in the industrial category.}
 \label{table:industrial}
 \end{table}
 
 \begin{table}[H]
 {\scriptsize
 \begin{center}
 \begin{tabular}{|l|c|c|c|c|c|c|c|c|}
 \hline
 \textbf{Solver} & \textbf{wcsp-dir} & \textbf{wcsp-log} & \textbf{HT} & \textbf{UP} & \textbf{PP} & \textbf{PWPMS} & \textbf{TT} & \textbf{Total} \\ \hline
 WMiFuMax        & 28.6\%            & 28.6\%            & 85\%        & 100\%       & 15.2\%      & 46\%           & 0\%         & 43.3\%         \\ \hline
 QWMaxSAT        & 66.7\%            & 61.9\%            & 20\%        & 0\%         & 40\%        & 17\%           & 30.8\%      & 34.5\%         \\ \hline
 Sat4j           & 14.3\%            & 14.3\%            & 15\%        & 37\%        & 38.7\%      & 2\%            & 30.8\%      & 21.7\%         \\ \hline
 MSUnCore        & 66.7\%            & 66.7\%            & 89\%        & 100\%       & 34.5\%      & 0\%            & 0\%         & 51\%           \\ \hline
 MaxSatz2013f    & 19\%              & 19\%              & 0\%         & 0\%         & 6.9\%       & 25\%           & 0\%         & 10\%           \\ \hline
 WMaxSatz-2009   & 19\%              & 14.3\%            & 0\%         & 41\%        & 5\%         & 12\%           & 0\%         & 13\%           \\ \hline
 WMaxSatz+       & 19\%              & 14.3\%            & 0\%         & 41\%        & 6.9\%       & 12\%           & 0\%         & 13\%           \\ \hline
 ISAC+           & 81\%              & 33.3\%            & 15\%        & 100\%       & 12.4\%      & 100\%          & 34.6\%      & 53.8\%         \\ \hline
 \end{tabular}
 \end{center}
 }
 \caption{Percentages of instances solved in the industrial category.}
 \end{table}

% = ROUND((b3/21)*100,1)
\begin{figure}[H]
\centering
\includegraphics[scale=0.53]{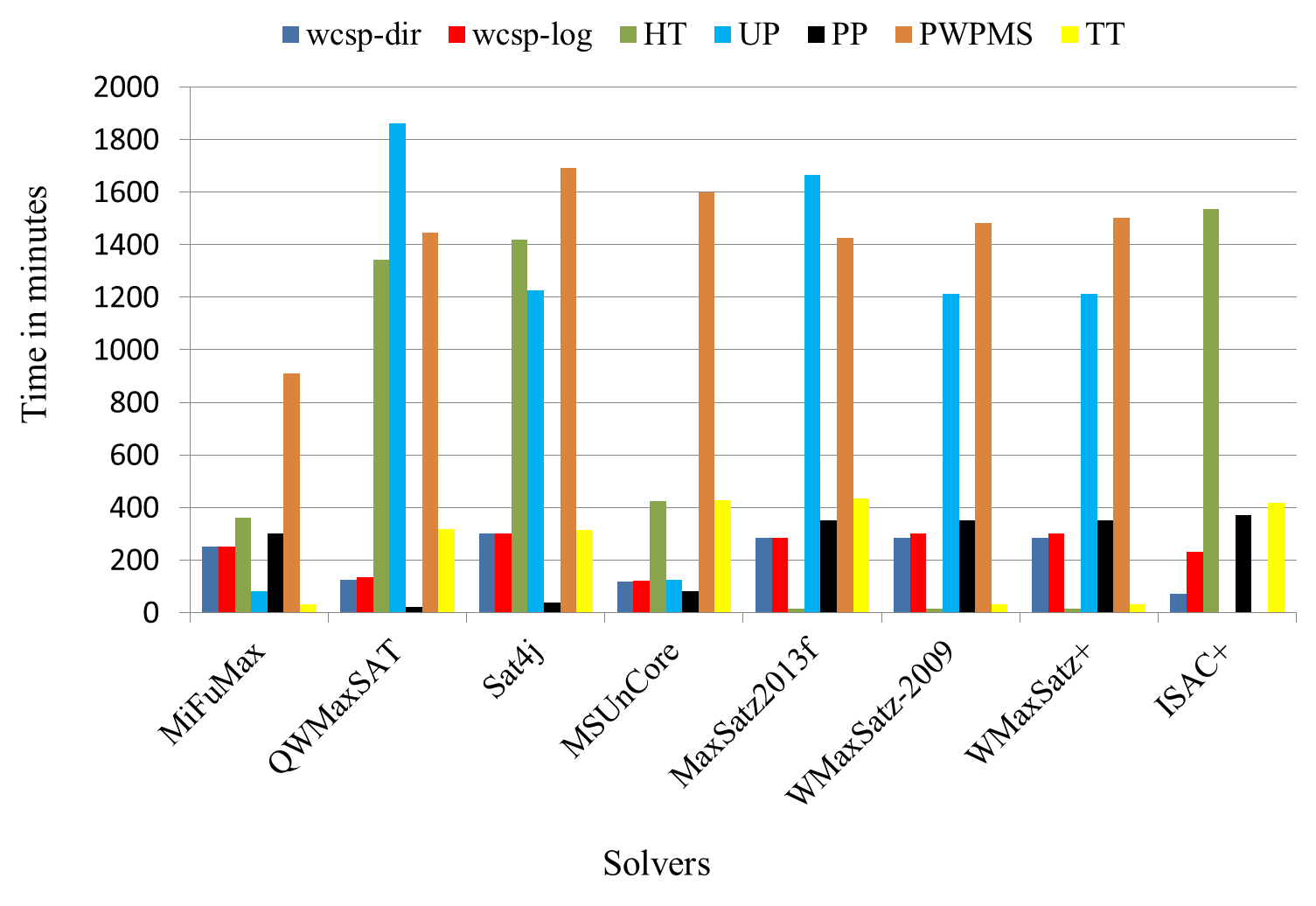}
\label{fig:random}
\caption{Time results for the industrial category.}
\label{fig:industrial1}
\end{figure}

It is clear that SAT-based solvers outperform branch and bound ones on industrial instances. The winner solver of this category in the 2013 MaxSAT evaluation is ISAC+ (see {\footnotesize \url{http://www.maxsat.udl.cat/13/results/index.html#wpms-industrial}}) and the same solver placed second in the 2014 evaluation (see {\footnotesize \url{http://www.maxsat.udl.cat/14/results/index.html#wpms-industrial-pc}}).

Generally, we can notice that on industrial instances, SAT-based solvers are performed considerably better than branch and bound solvers which performed poorly. On the other hand, branch and bound solvers outperformed SAT-based ones on random instances.

\section{Acknowledgments}
This paper is made possible through the help and support from Dr. Hassan Aly (Department of Mathematics, Cairo University, Egypt) and Dr. Rasha Shaheen (Department of Mathematics, Cairo University, Egypt). I would also like to thank Dr. Carlos Ans{\'o}tegui (University of Lleida, Spain) for his advice to include a section on translating pseudo Boolean constraints and his encouraging review of this work.

\bibliographystyle{plain}
\bibliography{references}

\end{document}